\documentclass[runningheads]{llncs}
\usepackage{graphicx}
\usepackage{comment}
\usepackage{amsmath,amssymb}
\usepackage{color}
\usepackage{url}
\usepackage{hyperref}

\newif\ifreview
\reviewfalse

\ifreview
	\usepackage{lineno}

	\linenumbers
\fi
\usepackage[T1]{fontenc}
\usepackage{esvect}
\usepackage{graphicx}
\usepackage{makecell}
\usepackage{multirow}
\usepackage{amsmath}
\usepackage{amssymb}
\usepackage{enumitem}
\usepackage{subcaption}
\usepackage{xcolor}
\usepackage[capitalize]{cleveref}
\usepackage{booktabs}

\newcommand{\ie}{\textit{i.e.}}
\newcommand{\eg}{\textit{e.g.}}

\begin{document}
\title{Road Obstacle Video Segmentation}
%
\def\SubNumber{105}

\def\GCPRTrack{Fast Track}
%


\ifreview
\else

	\author{Shyam Nandan Rai\inst{1} \and Shyamgopal Karthik \inst{2,3,4} \and Mariana-Iuliana Georgescu\inst{3,4} \and Barbara Caputo\inst{1} \and Carlo Masone\inst{1} \and Zeynep Akata\inst{3,4}}
	
	\authorrunning{S. Rai et al.}

    
	\institute{Politecnico di Torino \and University of Tübingen \and Technical University of Munich and Helmholtz Munich  \and  Munich Center for Machine Learning and Munich Data Science Institute}
\fi

\maketitle              
\begin{abstract}
With the growing deployment of autonomous driving agents, the detection and segmentation of road obstacles have become critical to ensure safe autonomous navigation. However, existing road-obstacle segmentation methods are applied on individual frames, overlooking the temporal nature of the problem, leading to inconsistent prediction maps between consecutive frames. In this work, we demonstrate that the road-obstacle segmentation task is inherently temporal, since the segmentation maps for consecutive frames are strongly correlated. To address this, we curate and adapt four evaluation benchmarks for road-obstacle video segmentation and evaluate 11 state-of-the-art image- and video-based segmentation methods on these benchmarks. Moreover, we introduce two strong baseline methods based on vision foundation models. Our approach establishes a new state-of-the-art in road-obstacle video segmentation for long-range video sequences, providing valuable insights and direction for future research.
\end{abstract}

\section{Introduction}
\label{sec:intro}

\begin{figure}[!t]
  \centering
   \includegraphics[width=1\linewidth, height=0.3\linewidth]{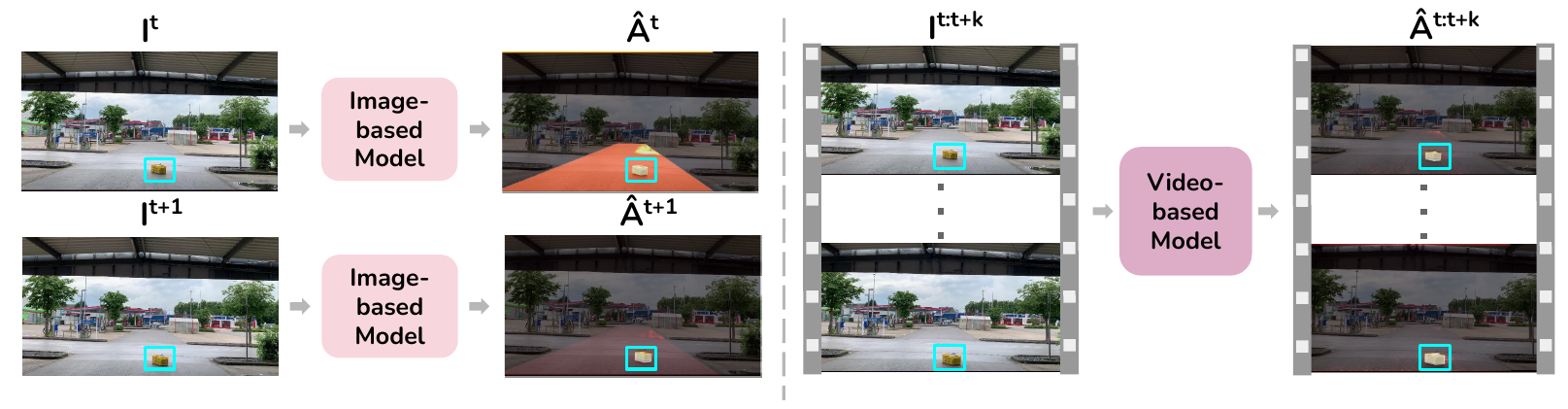}
   \caption{\textbf{Image- vs Video-based road-obstacle segmentation:} We illustrate the key difference between road-obstacle segmentation performance at image- and video-level on a set of consecutive image frames. Due to overlooking the temporal information, the image-level framework (left) fails to consistently predict the road-obstacle map across consecutive frames. However, the video-level framework (right) effectively leverages the temporal information, resulting in consistent maps. The box in blue encloses a road obstacle. }

   \label{fig:teaser}
\end{figure}

Semantic segmentation, \ie, the task of delineating objects in visual scenes with pixel-level granularity, is an essential component of the perception stack in autonomous cars~\cite{Cordts2016Cityscapes}. However, semantic segmentation models learn to identify and segment only objects that are contained in the training set. Such models, when deployed in the real world, could encounter a road-obstacle\footnote{Also referred to as \textit{anomaly segmentation} in the road-scene segmentation literature.} (novel object class absent during training), and incorrectly classify it as one of its known classes, potentially leading to unsafe behaviors. To address this safety-critical challenge, image-based segmentation methods~\cite{mask2anomaly,Maskomaly,RbA,grcic2020dense,EAM} have been extended to detect road obstacles (anomalies) in driving scenes. Currently, these models are evaluated on various benchmarks~\cite{chan2021segmentmeifyoucan,blum2021fishyscapes,lis2019detecting,lostandfound} consisting solely of static images, thus completely disregarding the inherent temporal consistency of the task. Consequently, existing image-based road-obstacle segmentation methods~\cite{mask2anomaly,Maskomaly,RbA,grcic2020dense,EAM} have inherent limitations, since they rely solely on individual frames (as shown in~\cref{fig:teaser}), despite the fact that driving applications utilize sensors that capture continuous data streams. Hence, incorporating temporal information is advantageous, as dependencies between consecutive frames enable models to leverage contextual cues from neighboring frames, leading to more accurate obstacle segmentation results.

Taking into account the limitations mentioned above, we propose using video-based methods to enhance the quality of road obstacle segmentation.  Hence, we introduce a new task of~\textbf{road-obstacle video segmentation}. This enables us to leverage temporal information, improving both the accuracy and consistency of road-obstacle segmentation models. To evaluate our proposed task, we establish benchmarks, evaluation metrics, and baseline methods. We collect relevant evaluation datasets for the task using the existing datasets, including Lidar-SOD~\cite{LiderSOD}, SOS~\cite{SOS}, and Lost\&Found~\cite{lostandfound}. However, these datasets lack long-term temporal sequences and are small in scale for the video road-obstacle segmentation task. To fill this gap, we create Apolloscape Road-Obstacle (AsRO), a large-scale road-obstacle dataset with long-range image sequences. We utilize the Apolloscape~\cite{ApolloScape} dataset to create AsRO. We employ existing evaluation metrics, such as AuROC and AuPRC, to assess the performance of image- and video-based methods. In particular, to evaluate the temporal consistency of the models, we employ the Video Consistency (VC)~\cite{VC} metric. However, we notice that VC does not penalize false positive pixels as it only takes into account the intersection of the predicted mask and the ground-truth mask. Therefore, we introduce $VC^{^*}$, a harmonic mean of video consistency for the road-obstacle and background classes that mitigates the aforementioned bias.

We benchmark these curated datasets on a set of 11 image- and video-based road-obstacle segmentation methods, reporting the performance in terms of image- and video-level metrics. Our key finding states that ignoring the temporal dimension leads to inconsistent road-obstacle predictions, as illustrated in~\cref{fig:teaser}. Furthermore, we introduce two strong road-obstacle segmentation baselines built on Mask2Former-Video~\cite{M2F-Video} and SAM 2~\cite{sam2}, which achieve state-of-the-art results on the proposed benchmarks. 
 
In summary, our contributions are the following.
\begin{itemize}
    \item We propose the road-obstacle (anomaly) video segmentation task, along with four novel benchmarks. 
    \item We evaluate, adapt, and retrain \text{11} state-of-the-art image- and video-based segmentation methods for the new setting on the proposed benchmarks. Moreover, we introduce the $VC^{^*}$ metric that penalizes the methods producing many false positive pixels.

   \item  We introduce two new baselines for road-obstacle video segmentation, leveraging foundation models. Our experiments indicate that these models originally developed for video segmentation can be effectively adapted for the similar-yet-distinct task of road-obstacle segmentation.

\end{itemize}

\section{Related Work}
\label{sec:related_work}

\noindent
\textbf{Semantic Segmentation.} Traditionally, semantic segmentation methods have been based on fully convolutional encoder-decoder architectures~\cite{long2015fully,chen2018encoder,lin2017refinenet,zhang2018exfuse}. However, there has been a paradigm shift towards transformer-based architectures~\cite{vaswani2017attention}, especially relying on the Swin transformer backbone~\cite{liu2021swin}. These methods~\cite{cheng2021per,cheng2022masked} introduce a mask transformer architecture and a per-pixel decoder to predict the segmentation output. More recently, there has also been the introduction of foundation segmentation models~\cite{sam,sam2} which tackle the segmentation task in a promptable manner.

\noindent
\textbf{Anomaly Segmentation.} Semantic segmentation models perform well on known classes but struggle to segment anomalies or unknown objects. Anomaly segmentation is the task of per-pixel labeling of image anomaly detection. This problem was first tackled by repurposing image-level methods to utilize softmax scores to predict anomalies~\cite{hendrycks2016baseline,liu2020energy,maxlogit,blum2021fishyscapes}. 
Afterwards, uncertainty quantification solutions have also been used for this task, including deep ensembles~\cite{fort2019deep}, Bayesian deep learning~\cite{mukhoti2018evaluating,gal2016dropout} and class-logits reasoning~\cite{jung2021standardized,hendrycks2022improving}. 
An alternative line of research has explored the usage of generative models trained on anomaly-free data to detect anomalies at testing times by their generation discrepancy~\cite{lis2019detecting,di2021pixel,xia2020synthesize,vojir2021road}.
Recently, a common practice has been to add supervision during training with outlier exposure to improve performance~\cite{bevandic2018discriminative,di2021pixel,ZhangSBHL23}. However, all these methods have typically been proposed to work on individual frames, without using the full temporal context provided in videos. Differently, we propose comprehensive video anomaly (road obstacles in our setup) segmentation benchmarks, and introduce novel methods based on the efficient adaptation of foundation segmentation models. 

\noindent \textbf{Video Anomaly Detection.} A different line of research focuses on video abnormal events detection~\cite{Ramachandra-WACV-2020,Ristea-CVPR-2024,Ahn_2024_ACCV}. In this scenario, an anomaly could be anything that diverges from the normal data through appearance or motion, even though the anomalous objects are included in the training data. Our work is different from this distantly related task, as our goal is to identify \textit{unseen} objects (obstacles) on the road.

   \noindent
\textbf{Anomaly Segmentation Benchmarks.} Several benchmarks have been proposed for anomaly segmentation in road scenes, such as FishyScapes~\cite{blum2021fishyscapes}, Road Anomaly~\cite{lis2019detecting}, Lost\&Found~\cite{lostandfound} and Segment Me If You Can~\cite{chan2021segmentmeifyoucan}. However, most of these datasets have been geared towards using static images. In this work, we establish video road-obstacle segmentation benchmarks from the Lidar SOD~\cite{LiderSOD}, Lost \& Found~\cite{lostandfound}, Apolloscape~\cite{ApolloScape}, and SOS~\cite{SOS} datasets. 

\section{Road-Obstacle Video Segmentation}
\label{sec:method}

The task of anomaly segmentation in urban road scenes as a video task is more suitable to achieve accurate and consistent anomaly segmentation. This is because an anomalous region detected in one frame is likely to appear in the following frames as well. Foundation models have become the standard approach for many computer vision tasks, suggesting their potential applicability to anomaly segmentation.

However, anomaly segmentation is not a conventional task, since it is highly dependent on the context and requires additional information. We investigate how a segmentation foundation model can be employed to segment anomalies in road scenes. Therefore, we propose two techniques based on foundation models.  The first strategy is to fine-tune the decoder of SAM 2 to achieve anomaly segmentation. The second technique is to employ a state-of-the-art segmentation model, namely M2FVideo~\cite{M2F-Video} and substitute its encoder with the powerful encoder of SAM 2~\cite{sam2}.  

\subsection{Problem Setting}
Given a segmentation model $\Psi$, our objective is to accurately segment road obstacles at the pixel level \textbf{throughout the video}. We train $\Psi$ using a set of video snippets and their corresponding \textit{semantic segmentation} ground-truth labels. It is important to note that during training, the video snippets do not contain any instances of road obstacles (anomalies). Consider a \textit{test} video sequence containing $T$ frames with road obstacles and a set temporal window that spans for $k$ frames. The $k$-span snippet from a random time stamp $t$, $I^{t:t+k} = \{ I^{t}, I^{t+1},\dots, I^{t+k}\}$, has the corresponding ground-truth road-obstacle snippet given as $A^{t:t+k} = \{ A^{t}, A^{t+1},\dots, A^{t+k}\}$. $A^{t}$ is a binary image map where $1$ denotes road obstacles and $0$ indicates background regions. During \textit{inference}, $\Psi$ functions as a road-obstacle video segmentation model by applying a simple approach of maximum softmax probability~\cite{hendrycks2016baseline}, predicting road-obstacle masks as $\hat{A}^{t:t+k} = \{ \hat{A}^{t}, \hat{A}^{t+1},\dots, \hat{A}^{t+k}\}$.

\begin{figure}[t]
  \centering
   \includegraphics[width=1\linewidth]{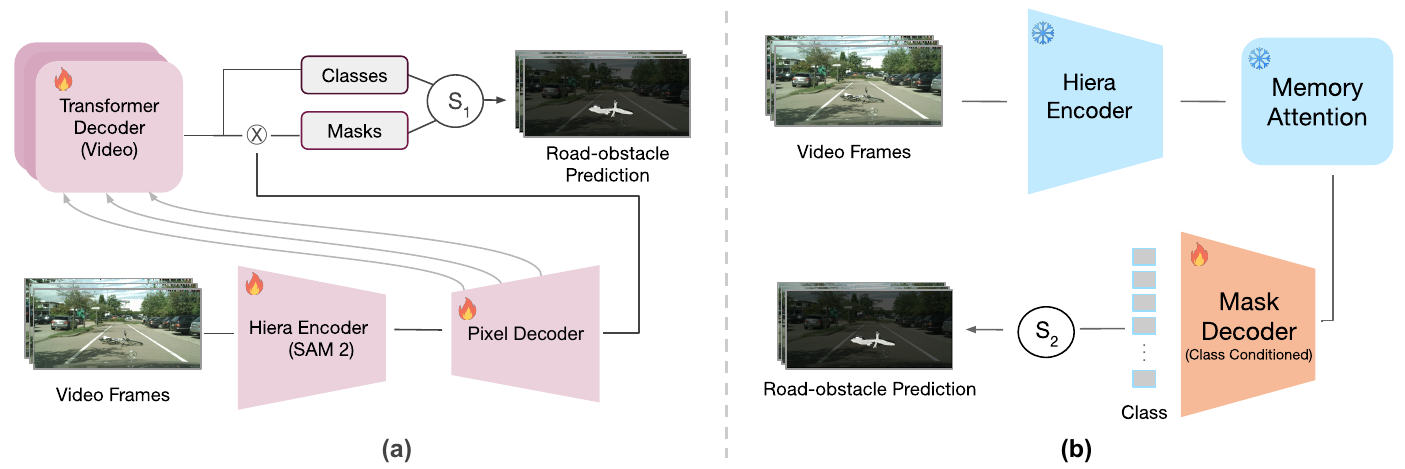}  
   \caption{(a) We present the \textbf{HM2F-Video} architecture. It consists of the Hiera backbone with the Mask2Former-video decoder. $S_1$ represents the road-obstacle scores calculated in~\cref{eq:HM2F}.(b) The \textbf{Class-Conditioned SAM 2} framework. We inject class information in SAM 2 by introducing a class-conditioned mask-decoder. $S_2$ represents the road-obstacle scores calculated in~\cref{Eq:CCSAM2}} 
   \label{fig:main_1-ab}
\end{figure} 

\subsection{Background}
We provide an overview of the Segment Anything Model 2 (SAM 2)~\cite{sam2} and Mask2Former-Video~\cite{mask2former} models, which serves as the basis for our methods.

\noindent
\textbf{SAM 2}~\cite{sam2} is a recent state-of-the-art model that extends SAM~\cite{sam}. SAM 2 is equipped with video processing, obtaining impressive results on multiple segmentation benchmarks~\cite{sam2}. Its architecture enabled large-scale training, resulting in the generation of high-quality class-agnostic segmentation masks during inference. SAM 2 architecture consists of: a) An \textit{image encoder} based on Hiera~\cite{Hiera} that was pre-trained using the MAE~\cite{MAE} technique. The Hiera architecture provides hierarchical features enabling decoding at multi-scale. b) A \textit{memory attention} that condition the image embeddings on the previous predictions and new prompts. c) A \textit{prompt encoder} which guides the model to discover potential objects in the input image, capable of processing points, bounding boxes, or masks. d) A \textit{memory encoder} that generates a memory by downsampling the output mask received from the mask decoder. e) A \textit{memory bank} that retains information about past predictions for the target objects in the video. f) A  \textit{mask decoder} that outputs the final predictions that are class-agnostic. 

\noindent\textbf{Mask2Former-Video}'s~\cite{M2F-Video} meta-architecture consists of three parts: a) a \textit{backbone} that acts as feature extractor for a set of video frames, b) a \textit{pixel-decoder} that upsamples the low-resolution features extracted from the backbone to get high-resolution \textit{per-pixel embeddings} of the frames, and c) a \textit{video transformer decoder} based on Mask2Former~\cite{mask2former} with an additional joint spatio-temporal masked attention and a temporal positional encoding.

\subsection{HM2F-Video}
SAM 2~\cite{sam2} is trained on millions of masks, providing an effective image encoder with generalized features. Therefore, we utilize the Hiera backbone~\cite{Hiera} together with the Mask2Former-video~\cite{M2F-Video} decoder. We name this method HM2F-Video and illustrate it in~\cref{fig:main_1-ab}(a). During inference, the model predicts a set of \textit{class masks} $M\in \mathbb{R}^{N\times \tau\times (H \times W)}$ and their associated \textit{class scores} $C\in \mathbb{R}^{N \times K}$, where $N$ is the number of queries, $K$ is the number of training classes, $\tau$ is the number of frames, and $H$ and $W$ are the spatial dimensions. The road-obstacle segmentation score for an input video $I$ is computed as:
\begin{equation}
    \label{eq:HM2F}
    S_{1}(I) = 1 - \max_{k=1}^{K} \left(C^T \times M\right).
\end{equation}

\subsection{Class Conditioned SAM2}
We introduce Class-Conditioned SAM 2 (CC-SAM2) that injects class information into SAM 2, by adding a class-conditioned mask decoder into the SAM 2 architecture, as illustrated in~\cref{fig:main_1-ab}(b). The class-conditioned mask decoder architecture is similar to the SAM 2 class-agnostic mask decoder, but differs in the number of output masks, which depends on the number of classes contained in the training dataset. During training, the SAM 2 model is frozen, while only the class-conditioned mask decoder is updated. During inference, the road-obstacle segmentation score of an input video $I$ is computed as:
\begin{equation}
    \label{Eq:CCSAM2}
    S_2(I) = -\max_{k=1}^{K}\Upsilon(I),
\end{equation}

\noindent
where $S_2$ and $\Upsilon$ are the scoring function and the CC-SAM2 model, respectively. 

\begin{table}[!t]
\centering
\renewcommand{\arraystretch}{1}
\setlength\tabcolsep{4pt}
\caption{\textbf{Dataset Statistics.} We compare our proposed Apolloscape-Road-Obstacle (AsRO) with SOS~\cite{SOS}, Lost\&Found~\cite{lostandfound}, and LidarSOD~\cite{LiderSOD}. Highest value in each column is bold.}

{%
\begin{tabular}{lcccc}
\toprule
\multirow{1}{*}{\textbf{Dataset}} & \multirow{1}{*}{\textbf{Resolution}} & \multirow{1}{*}{\textbf{\#Frames}} & \textbf{Average Length} & \multirow{1}{*}{\textbf{Location}} \\
\midrule
SOS~\cite{SOS} & 1920$\times$1080 & 1004 & 56.4 & Germany \\
Lost\&Found~\cite{lostandfound} & 2048$\times$1024 & 1779 & 149.2 & Germany \\
LidarSOD~\cite{LiderSOD} & 1280$\times$720 & 2421 & 194.6 & India \\
\midrule
\textbf{AsRO (Ours)} & \textbf{3384$\times$2710} & \textbf{12480} & \textbf{1192.7} & China \\
\bottomrule
\end{tabular}%
}
\vspace{-1em}
\label{tab:dataset_stats}
\end{table}

\section{Road-Obstacle Video Segmentation Benchmark}

There are several benchmarks for image-based road-obstacle (anomaly) segmentation~\cite{blum2021fishyscapes,chan2021segmentmeifyoucan}\footnote{Also refer to as anomaly segmentation in the road-scene segmentation literature.}. However, these benchmarks are not suitable for the road-obstacle video segmentation task, because they lack temporal consistency and evaluation protocols to assess model performance for videos. Therefore, we adapt the existing road-scene segmentation datasets, namely SOS~\cite{SOS}, LidarSOD~\cite{LiderSOD}, and Lost\&Found~\cite{lostandfound} to road-obstacle segmentation. To create the datasets, we only select the obstacles that are located on the road. However, the aforementioned datasets do not have long-range temporal sequences or the scale to assess this task effectively. Therefore, we propose Apolloscape-Road-Obstacle (AsRO), a large-scale road-obstacle video dataset consisting of long-range videos.

\begin{figure}[t]
  \centering
   \includegraphics[width=0.75\linewidth, height=0.35\linewidth]{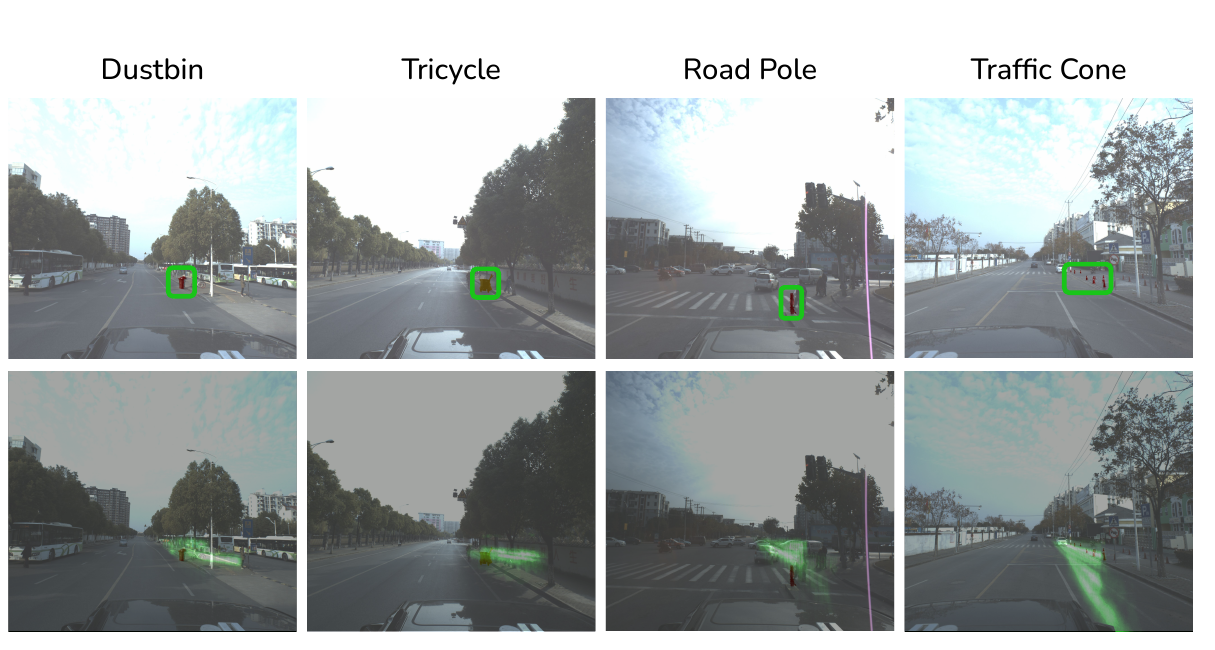}
   \caption{\textbf{Apolloscape-Road-Obstacle}:
   \textit{Top:} Sample images of each anomaly type.~\textit{Bottom:} Average location per anomaly type. Please zoom in for clarity.}
   \label{fig:apolloscapes_samples}
\end{figure}

\subsection{Apolloscape-Road-Obstacle (AsRO):} 

We curate the Apolloscape dataset~\cite{ApolloScape}, which consists of 143K video frames. Among all frames, 51,865 frames contain obstacles on the road labeled as: Traffic cone, Road pole, Tricycle, Tricycle group and Dustbin. However, we observe that some images have obstacles outside the road. Therefore, we remove all obstacles that do not have their center on the road. After this step, we obtain 12,480 images in 12 video sequences, with an average sequence length of 1192 frames.~\Cref{tab:dataset_stats} presents the comparison between AsRO and the other datasets. We notice that our AsRO dataset has the highest number of frames and the highest average sequence length among the compared datasets.

\subsection{Evaluation Metrics} 
We provide a holistic evaluation of the road-obstacle video segmentation methods by employing the following metrics.

\noindent
{\textbf{Pixel-level.}} We consider the Area under the Receiver Operating Curve (AuROC), the Area under the Precision-Recall Curve
(AuPRC), and the False Positive Rate at a True Positive Rate of 95\% (FPR$_{95}$). \\

\noindent
{\textbf{Component-level.}} We employ component-level metrics, as they are not affected by the size of the object, unlike pixel-level metrics, which are biased toward object size. We use an adjusted version of the component-wise intersection over union (sIoU) that excludes pixels that correctly intersect with another ground-truth component, positive predictive value (PPV component-wise precision), and component-wise F1-score (F1$_1^{*}$).

\noindent{\textbf{Video-level.}} We employ Video Consistency~\cite{VC}, a video evaluation metric that is widely used to assess the consistency of segmentation predictions. Given a ground-truth binary mask $\{A_{i}\}_{i=1}^{T}$ and a predicted binary mask $\{\hat{A}_{i}\}_{i=1}^{T}$, where $T$ is the number of frames, the video consistency is computed as:

\vspace{-1.em}
\begin{equation}
\label{VC}
VC_T = \frac{(\hat{A}_{1} \cap \ldots \cap \hat{A}_{T}) \cap (A_{1} \cap \ldots \cap A_{T})}{(A_{1} \cap \ldots \cap A_{T})}.
\end{equation}

Calculating the VC score only for the road-obstacle class gives an advantage to methods that have a higher false positive rate since the intersection between the ground truth maps and the prediction maps would be higher.
To alleviate this problem, we introduce $VC^{*}$, a harmonic mean of video consistency for road-obstacle $VC^{RO}$, and background (non-road obstacle area) $VC^{BG}$ given as:
\begin{equation}
\label{VC_ours}
VC^{*} = 2 \frac{VC^{RO}\cdot VC^{BG}}{VC^{RO}+VC^{BG}}.
\end{equation}

We select the harmonic mean because it achieves a trade-off between the true positive and the false negative rates, similar to the well-known F1-score. For example, a perfect video segmentation model would obtain a maximum score for $VC^{RO}$ and $VC^{BG}$  equal to 1, therefore the value of $VC^{*}$ would be 1. On the other hand, this metric effectively penalizes methods that predict many false positives, which also have high $VC^{RO}$ scores (many false positive obstacles lead to a higher probability of matching the ground truth maps), but the $VC^{BG}$ score would be low, because the background would not be segmented correctly, leading to a lower $VC^{*}$ score.

\subsection{Baselines}
We establish two baseline categories: (a) \textbf{Image-based} obstacle segmentation methods: Entropy~\cite{hendrycks2016baseline}, Max Softmax~\cite{hendrycks2016baseline}, Energy~\cite{liu2020energy}, Max Logit~\cite{maxlogit}, Void Classifier~\cite{blum2021fishyscapes}, Mask2Anomaly~\cite{rai2023unmasking}, EAM~\cite{EAM}, AEM~\cite{EAM}, and RbA~\cite{RbA}; (b) \textbf{Video-based} semantic segmentation methods: M2F-Video~\cite{M2F-Video} and DVIS~\cite{DVIS}, adapted to our task by Max Softmax~\cite{hendrycks2016baseline}. We evaluate all baselines using the metrics and datasets and compare them with our proposed methods in the following~sections.

\begin{table*}[t]
\centering
\setlength{\tabcolsep}{10pt} 
\caption{Quantitative results on \textbf{Lost\&Found}~\cite{lostandfound}. We compare our proposed baselines, CC-SAM2 and HM2F-Video, to various state-of-the-art methods~\cite{hendrycks2016baseline,liu2020energy,maxlogit,blum2021fishyscapes,rai2023unmasking,EAM,RbA,M2F-Video,DVIS}.
Best results are in \textbf{bold} and second best are \underline{underlined}. }

\resizebox{0.99\linewidth}{!}{\begin{tabular}{@{}llccccccc@{}}
\toprule
& \multirow{2}{*}{\bf Methods} & \multicolumn{3}{c}{\bf Pixel-Level Metrics} & \multicolumn{3}{c}{ \bf Component-Level Metrics} & \multicolumn{1}{c}{ \bf Video Metrics} \\
\cmidrule(lr){3-5} \cmidrule(lr){6-8} \cmidrule(lr){9-9} 
& & AuROC$\uparrow$ & AuPRC$\uparrow$ & FPR$_{95}\downarrow$ & sIoU$\uparrow$ & PPV$\uparrow$ & F1$^{*}_{1}\uparrow$ & VC$^{*}\uparrow$\\
\midrule
\multirow{9}{*}{\rotatebox[origin=c]{90}{Image Level}}
& Random & 50.00 & 0.73 & 94.99 & 0.53 & 1.33 & 0.00 &  0.00\\
& Entropy~\cite{hendrycks2016baseline} & 86.72 & 5.65 & 72.32 & 28.83 & 26.63 & 26.70 &22.87\\
& Energy~\cite{liu2020energy} & 85.88 & 6.22 & 78.69 & 28.59 & 25.92 & 25.70  &0.00\\
& Max Softmax~\cite{hendrycks2016baseline} & 89.03 & 7.31 & 56.08 & 37.55 & 35.48 & 35.57 & 22.84\\
& Max Logit~\cite{maxlogit} & 90.53 & 10.39 & 36.08 & \textbf{46.10} & 42.73 & 44.34  &\underline{52.11}\\
& Void Classifier~\cite{blum2021fishyscapes} & 54.43 & 11.28 & 99.99 & 0.54 & 1.80 & 0.05 &23.34\\
& Mask2Anomaly~\cite{rai2023unmasking} & 92.95 & 14.99 & \underline{22.81} & 39.34 & 47.03 & 41.29 &48.86\\
& EAM~\cite{EAM} & 84.20 & 7.33 & 93.27 & 41.11 & \underline{53.42} & \textbf{46.67}& 49.33 \\
& AEM~\cite{EAM} & 83.89 & 7.15 & 92.80 & 41.02 & \textbf{53.48} & \underline{46.57} &49.22\\
& RbA~\cite{RbA} & 87.31 & 43.01 & 94.40 & 7.54 & 9.32 & 7.53 &36.54 \\
 \midrule
\multirow{4}{*}{\rotatebox[origin=c]{90}{Video Level}}
& M2F-Video~\cite{M2F-Video} & \underline{95.11} & \textbf{60.51} & 27.31 & 38.33 & 50.59 & 37.25 & 35.06  \\
& DVIS~\cite{DVIS} & 93.33 & \underline{59.91} & 57.59 & \underline{43.51} & 29.66 & 30.78 &25.24  \\
& CC-SAM2 (Ours) & \textbf{98.44} & 57.83 & \textbf{6.22} & 15.13 & 38.90 & 14.70  &\textbf{54.15}\\
& HM2F-Video (Ours) & 93.87 & 54.00 & 53.59 & 31.74 & 50.92 & 34.67  & 32.43\\
\addlinespace[0.3em]
\bottomrule
\end{tabular}}
\label{tab:lostandfound}
\end{table*}

\section{Experiments}
\subsection{Implementation Details}
\noindent Implementation details about the baselines are in the supplementary.  

\noindent
\textbf{HM2F-Video:} The image encoder is the pre-trained Hiera-B+ from SAM 2 and the decoder architecture is adopted from Mask2Former-Video. During training, two consecutive frames were processed setting the batch size to $4$. We trained the model for $32,\!000$ iterations while keeping the learning rate equal to $1e\!-\!4$. The number of mask-decoder layers was set to $9$, while the number of output queries is equal to $100$.

\noindent
\textbf{CC-SAM2:} In CC-SAM2, we utilized SAM 2 with Hiera-B+ as image encoder. The model was trained for 40 epochs with a batch size of $1$. The base learning rate is set to $1e\!-\!4$, and the input image resolution is 1024$\times$1024 pixels. The number of consecutive input video frames is set to $2$.

\begin{table*}[t]
\centering
\setlength{\tabcolsep}{10pt}
\caption{Quantitative results on \textbf{SOS}~\cite{SOS}.  We compare our proposed baselines, CC-SAM2 and HM2F-Video, to various state-of-the-art methods~\cite{hendrycks2016baseline,liu2020energy,maxlogit,blum2021fishyscapes,rai2023unmasking,EAM,RbA,M2F-Video,DVIS}. Best results are in \textbf{bold} and second best are \underline{underlined}.}

\resizebox{0.99\linewidth}{!}{\begin{tabular}{@{}llccccccc@{}}
\toprule
& \multirow{2}{*}{\bf Methods} & \multicolumn{3}{c}{\bf Pixel-Level Metrics} & \multicolumn{3}{c}{\bf Component-Level Metrics} & \multicolumn{1}{c}{\bf Video Metrics} \\
\cmidrule(lr){3-5} \cmidrule(lr){6-8} \cmidrule(lr){9-9} 
& & AuROC$\uparrow$ & AuPRC$\uparrow$ & FPR$_{95}\downarrow$ & sIoU$\uparrow$ & PPV$\uparrow$ & F1$^{*}_{1}\uparrow$ & VC$^{*}\uparrow$\\
\midrule
\multirow{10}{*}{\rotatebox[origin=c]{90}{Image Level}}
& Random & 50.00 & 0.89 & 95.00 & 0.73 & 1.76 & 0.00 &0.00  \\
& Entropy~\cite{hendrycks2016baseline} & 96.43 & 39.16 & 20.50 & 31.24 & 12.91 & 17.47 &13.31\\
& Energy~\cite{liu2020energy} & 95.20 & 65.85 & 24.72 & 27.53 & 11.46 & 15.57 &0.00\\
& Max Softmax~\cite{hendrycks2016baseline} & 96.99 & 72.24 & 20.05 & 44.23 & 20.26 & 25.90 &13.32\\
& Max Logit~\cite{maxlogit} & 96.94 & 74.44 & 20.99 & \textbf{53.42} & 24.14 & \underline{31.35} &\textbf{60.21}  \\
& Void Classifier~\cite{blum2021fishyscapes} & 68.39 & 41.81 & 99.96 & 0.65 & 3.17 & 0.01  &21.33\\
& Mask2Anomaly~\cite{rai2023unmasking} & 96.10 & 64.81 & 25.11 & \underline{45.35} & \underline{30.24} & \textbf{31.97}  &\underline{52.42}\\
& EAM~\cite{EAM} & 90.32 & 71.76 & 64.41 & 41.95 & 24.44 & 28.76 &43.75 \\
& AEM~\cite{EAM} & 90.22 & 71.35 & 64.03 & 41.96 & 23.64 & 28.08  &43.79\\
& RbA~\cite{RbA} & 97.09 & \textbf{86.19} & \underline{7.67} & 8.20 & 9.31 & 7.47 &44.70 \\

\midrule
\multirow{4}{*}{\rotatebox[origin=c]{90}{Video Level}}
& M2F-Video~\cite{M2F-Video} &  \underline{97.41} & 57.23 & 15.61 & 48.71 & 19.43 & 21.05 &43.13 \\
& DVIS~\cite{DVIS} & 94.34 & 14.22 & 28.65 & 45.17 & 9.50 & 10.89 &26.85  \\

& CC-SAM2 (Ours) & \textbf{99.24} & \underline{79.24} & \textbf{3.69} & 31.14 & \textbf{47.00} & 28.14 &43.27  \\
& HM2F-Video (Ours) & 83.62 & 1.23 & 46.03 & 8.35 & 6.96 & 4.67 &7.93 \\
\addlinespace[0.3em]
\bottomrule
\end{tabular}}
\label{tab:sos}
\end{table*}

\begin{table*}[t]
\centering
\setlength{\tabcolsep}{10pt}
\caption{Quantitative results on \textbf{Lidar SOD}~\cite{LiderSOD}.  We compare our proposed baselines, CC-SAM2 and HM2F-Video, to various state-of-the-art methods~\cite{hendrycks2016baseline,liu2020energy,maxlogit,blum2021fishyscapes,rai2023unmasking,EAM,RbA,M2F-Video,DVIS}. Best results are in \textbf{bold} and second best are \underline{underlined}.}

\resizebox{0.99\linewidth}{!}{\begin{tabular}{@{}llccccccc@{}}
\toprule
& \multirow{2}{*}{\bf Methods} & \multicolumn{3}{c}{\bf Pixel-Level Metrics} & \multicolumn{3}{c}{\bf Component-Level Metrics} & \multicolumn{1}{c}{\bf Video Metrics} \\
\cmidrule(lr){3-5} \cmidrule(lr){6-8} \cmidrule(lr){9-9} 
& & AuROC$\uparrow$ & AuPRC$\uparrow$ & FPR$_{95}\downarrow$ & sIoU$\uparrow$ & PPV$\uparrow$ & F1$^{*}_{1}\uparrow$ & VC$^{*}\uparrow$\\
\midrule
\multirow{10}{*}{\rotatebox[origin=c]{90}{Image Level}}
& Random & 50.00 & 0.39 & 94.99 & 0.14 & 0.64 & 0.01  &0.00 \\
& Entropy~\cite{hendrycks2016baseline} & 87.21 & 2.03 & 55.60 & 18.17 & 11.93 & 13.99 &3.66  \\
& Energy~\cite{liu2020energy} & 85.15 & 1.77 & 75.17 & 14.92 & 10.47 & 12.01 &0.00 \\
& Max Softmax~\cite{hendrycks2016baseline} & 88.29 & 1.84 & 27.06 & 21.86 & 11.99 & 14.02 &3.65\\
& Max Logit~\cite{maxlogit} & 88.67 & 1.80 & \underline{26.14} & \textbf{23.97} & 12.82 & 14.56 &37.53\\
& Void Classifier~\cite{blum2021fishyscapes} & 50.45 & 0.71 & 99.80 & 0.12 & 0.69 & 0.02 &10.53\\
& Mask2Anomaly~\cite{rai2023unmasking} & 84.39 & 2.55 & 46.42 & 14.69 & 9.06 & 7.63 &26.36\\
& EAM~\cite{EAM} & \underline{91.40} & \underline{5.10} & 30.91 & \underline{22.60} & \underline{14.02} & \textbf{14.90} & \underline{37.81} \\
& AEM~\cite{EAM} & 91.17 & 4.89 & 31.36 & 22.42 & \textbf{14.06} & \underline{14.82} &37.61 \\
& RbA~\cite{RbA} & 86.74 & 2.05 & 86.42 & 0.89 & 1.80 & 0.95  &23.99\\
\midrule
\multirow{4}{*}{\rotatebox[origin=c]{90}{Video Level}}
& M2F-Video~\cite{M2F-Video} & 81.43 & 1.09 & 54.09 & 9.63 & 10.39 & 6.08 &20.37 \\
& DVIS~\cite{DVIS} & 79.58 & 1.00 & 61.82 & 12.84 & 8.63 & 6.53 &15.84  \\
& CC-SAM2 (Ours) & \textbf{92.52} & \textbf{14.00} & \textbf{25.56} & 9.11 & 4.96 & 3.25 &\textbf{40.48}  \\
& HM2F-Video (Ours) & 83.62 & 1.23 & 46.03 & 8.35 & 6.96 & 4.67& 13.80  \\
\addlinespace[0.3em]
\bottomrule
\end{tabular}}
\label{tab:lidersod}
\end{table*}

\begin{figure}[h!]
  \centering
   \includegraphics[width=1.0\linewidth]{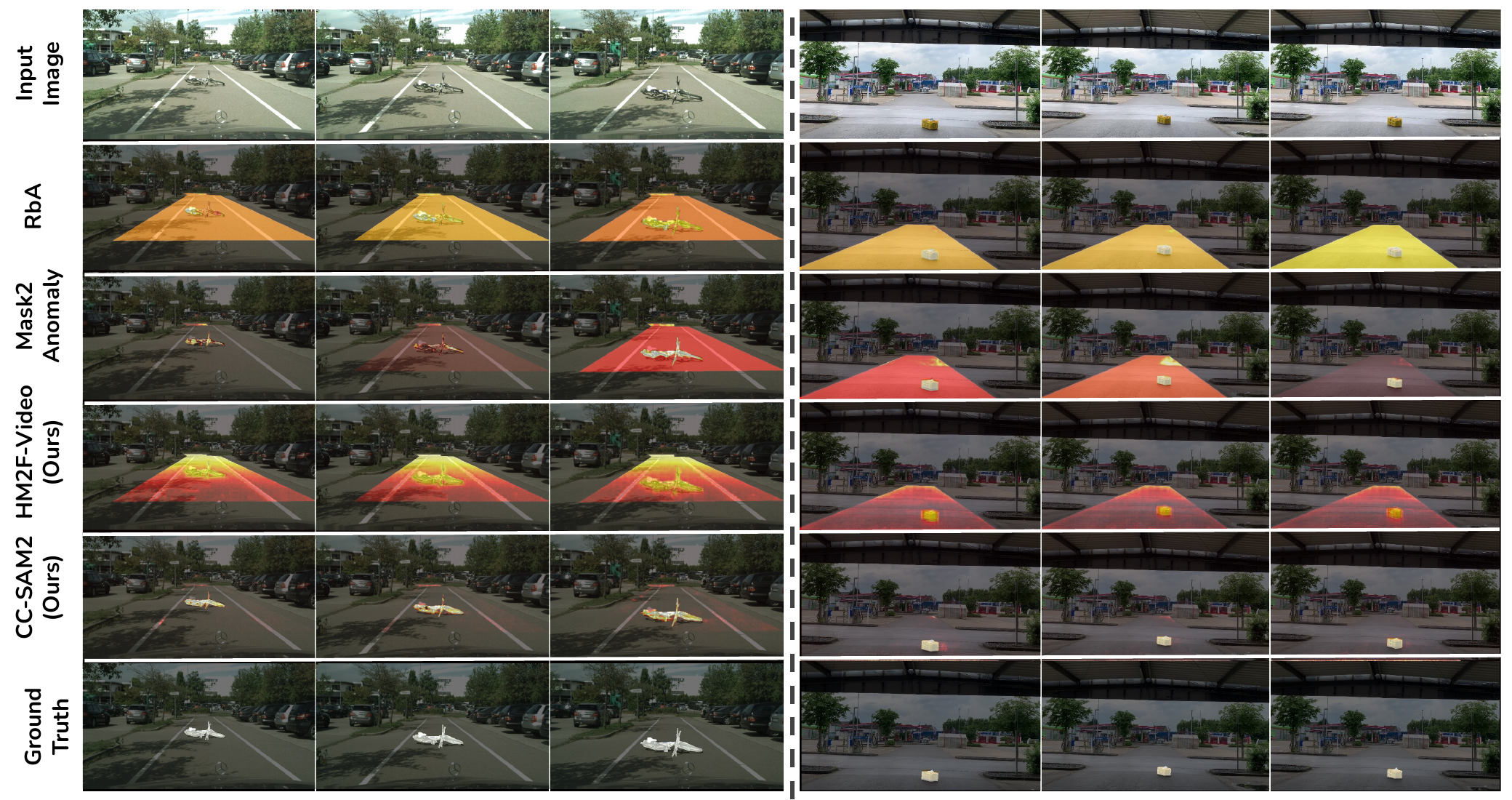}

   \caption{\textbf{Qualitative Results}. Road-obstacle prediction maps from RbA~\cite{RbA}, Mask2Anomaly~\cite{mask2anomaly}, HM2F-Video and CC-SAM2 on the SOS dataset~\cite{SOS}.  HM2F-Video accurately segments road-obstacles across temporal frames.} 

   \label{fig:qualitative}
\end{figure}

\begin{table}[t]
\centering
\setlength{\tabcolsep}{10pt}
\caption{Quantitative results on Apolloscape-Road-Obstacle. Best results are in \textbf{bold} and second best are \underline{underlined}. ${\dag}$ are video-based road obstacle segmentation methods.}

\resizebox{0.99\linewidth}{!}{\begin{tabular}{@{}llccccccc@{}}
\toprule
& \multirow{2}{*}{\bf Methods} & \multicolumn{3}{c}{\bf Pixel-Level Metrics} & \multicolumn{3}{c}{\bf Component-Level Metrics} & \multicolumn{1}{c}{\multirow{1}{*}{\bf Video Metrics}} \\
\cmidrule(lr){3-5} \cmidrule(lr){6-8} \cmidrule(lr){9-9}
& & AuROC$\uparrow$ & AuPRC$\uparrow$ & FPR$_{95}\downarrow$ & sIoU$\uparrow$ & PPV$\uparrow$ & F1$^{*}_{1}\uparrow$ & VC$^{*}$ $\uparrow$ \\
\midrule
& Random & 49.92 & 0.02 & 95.09 & 4.1 & 0.23 & 0.04 & 0.00 \\
& Mask2Anomaly~\cite{rai2023unmasking} & 89.94 & 0.44 & 64.62 & {10.01} & 0.35 & 0.38 &  0.02\\
& EAM~\cite{EAM} & 64.67 & \underline{0.72} & 99.71 & 5.43 & \textbf{0.73} & \underline{0.47} & 1.62 \\
& M2F-Video~\cite{M2F-Video}~$^{\dag}$ & 92.66 & 0.50 & 29.81 & 8.91 & \underline{0.59} & 0.44 & 1.85 \\
& DVIS~\cite{DVIS}~$^{\dag}$ & \underline{94.07} & 0.68 & \underline{21.69} & \underline{11.41} & 0.52 & \underline{0.47} & 1.03 \\
& CC-SAM2 (Ours)~$^{\dag}$ & \textbf{98.83} & \textbf{2.13} & \textbf{3.90} & \textbf{16.79} & 0.48 & \textbf{0.48} & \textbf{8.28} \\
\bottomrule
\end{tabular}}
\vspace{-1.5em}
\label{tab:apollo}
\end{table}

\subsection{Results}
\textbf{Pixel-Level.} We report the pixel-level evaluation details on the SOS~\cite{SOS}, LidarSOD~\cite{LiderSOD}, Lost\&Found~\cite{lostandfound} and AsRO datasets in~\cref{tab:lostandfound,tab:sos,tab:lidersod,tab:apollo}. We observe the FPR$_{95}$ performance obtained by CC-SAM2 of 3.69\% (SOS), 25.56\% (LidarSOD), 3.90\% (AsRO), and 6.22\% (Lost\&Found), is the lowest among all compared methods. This demonstrates that leveraging temporal information in video-level methods significantly reduces false positives. Moreover, video-level methods also show strong performance on other pixel-level evaluation metrics.\\

\noindent\textbf{Component-Level. }~\Cref{tab:lostandfound,tab:sos,tab:lidersod,tab:apollo}  present the results obtained on the road-obstacle video segmentation datasets in terms of component-level metrics.  Notably, the image-based models reach the highest performance, for instance EAM obtains an F1$^{*}_{1}$ value
of 46.67\%~(Lost\&Found), AEM reaches a PPV score  of 14.06\%~(Lidar SOD), and MaxLogit attains an sIoU value of 53.42\%~(SOS). In contrast, on the AsRO benchmark, CC-SAM2 stands out as one of the best performing methods, achieving a score of 16.79\% in terms of sIoU and an F1$^{*}_{1}$ value of 0.48\%. These results reinforce the observation that image-level methods are strong baselines. However, for the road-obstacle video segmentation task, video-based methods perform better longer video sequence dataset i.e. AsRO. 

\noindent\textbf{Video-Level. } In ~\Cref{tab:lostandfound,tab:sos,tab:lidersod,tab:apollo}, we report the evaluation in terms of video-level metrics on the road-obstacle video segmentation datasets. CC-SAM2 achieves strong performance in terms of VC$^{*}$ across all datasets. In particular, on the AsRO dataset, CC-SAM2 attains a VC$^{*}$ score of 8.28, which is significantly better than the baseline methods (the second-best score of 1.85 is reached by M2F-Video). This highlights the effectiveness of video-based approaches for road-obstacle video segmentation.

\noindent
\textbf{Qualitative Results.}~\cref{fig:qualitative} presents a qualitative comparison between the baselines, RbA~\cite{RbA} and Mask2Anomaly~\cite{mask2anomaly}, and our proposed HM2F-Video and CC-SAM2. We notice that CC-SAM2 provides coherent and accurate road-obstacle segmentation maps, outperforming the image-based methods. In contrast, Mask2Anomaly~\cite{mask2anomaly} and RbA~\cite{RbA} predict many false positives, indicating the importance of temporal consistency.

\label{sec:experiments}
\subsection{Discussion}
\textbf{Image-based vs Video-based methods:} We observe from~\cref{tab:lidersod,tab:lostandfound,tab:sos}, that the performance of image-based methods are close to the video-based ones, in terms of pixel-level metrics (EAM~\cite{EAM} obtains the second-best score on LidarSOS~\cite{LiderSOD},~\cref{tab:sos}). In terms of output consistency (VC), most image-based methods obtain low scores (RbA~\cite{RbA} attains 36.54\% on Lost\&Found~\cite{lostandfound},~\cref{tab:lostandfound}), showing limited correlation between the output of consecutive frames. However, as shown in~\cref{tab:apollo}, CC-SAM2 achieves the best performance on almost all evaluation metrics. In summary, our empirical experiments indicate that image-based methods perform better on short video sequences, whereas for longer sequences i.e AsRO dataset, video-based methods like CC-SAM2 show better results.

\noindent\textbf{CC-SAM2 and HM2F-Video Performance:} From~\cref{tab:lidersod,tab:lostandfound,tab:sos}, we observe a significant performance gap between CC-SAM2 and HM2F-Video, which is attributed to differences in their architectural designs. CC-SAM2 is based on SAM2, a foundation model architecture showing strong generalizability, whereas HM2F-Video is based on M2F-Video architecture\cite{M2F-Video}.

\begin{table}[t]
\renewcommand{\arraystretch}{1}
\caption{Ablation results performance obtained on the SOS dataset~\cite{SOS}. Top results are highlighted in \textbf{bold}.}  
\label{tab:abl2}
\begin{subtable}{0.99\linewidth}
        \caption{Performance of the HM2F-Video method when the encoder is frozen or finetuned.}\label{tab:abl_frozen}
        \centering
        \resizebox{0.75\linewidth}{!}{
        \begin{tabular}{@{}llcccccc@{}}
        \toprule
         & \multirow{2}{*}{\bf Encoder} & \multicolumn{3}{c}{\bf Pixel-Level Metrics} & \multicolumn{3}{c}{\bf Component-Level Metrics} \\
        \cmidrule(lr){3-5} \cmidrule(lr){6-8}
         &  & AuROC$\uparrow$ & AuPRC$\uparrow$ & FPR$_{95}\downarrow$ & sIoU$\uparrow$ & PPV$\uparrow$ & F1$^{*}_{1}\uparrow$ \\
        \midrule
        & Freeze & 77.79 & 3.30 & 77.60 & 33.00 & 2.28 & 3.10 \\
        & Unfreeze & \textbf{93.44} & \textbf{53.89} & \textbf{43.38} & \textbf{46.96} & \textbf{7.15} & \textbf{8.87} \\
        \bottomrule
        \end{tabular}
        } 
\end{subtable}
 \begin{subtable}{0.99\linewidth}
         \caption{CC-SAM2 performance when varying the number of input frames.}\label{tab:abl_frames}
        \centering
        \resizebox{0.75\linewidth}{!}{\begin{tabular}{@{}lccccccc@{}}
            \toprule
            & \multirow{2}{*}{\bf \#Frames} & \multicolumn{3}{c}{\bf Pixel-Level Metrics} & \multicolumn{3}{c}{\bf Component-Level Metrics} \\
            \cmidrule(lr){3-5} \cmidrule(lr){6-8}
            & & AuROC$\uparrow$ & AuPRC$\uparrow$ & FPR$_{95}\downarrow$ & sIoU$\uparrow$ & PPV$\uparrow$ & F1$^{*}_{1}\uparrow$ \\
            \midrule
            &2    &\textbf{99.16}    &\textbf{65.29}    &3.48    &\textbf{31.92}    &47.44    &\textbf{26.26}\\
            &4    &99.15    &42.63    &\textbf{2.71}    &22.44    &\textbf{59.63}     &25.64 \\
            &8    &97.71    &16.92    &6.13    &3.89     &26.47     &2.71 \\
            
            \bottomrule
            \end{tabular}}
            \vspace{-1.0em}
    \end{subtable}

\end{table}

\subsection{Ablation Results}
We utilize SOS~\cite{SOS} as the evaluation dataset for ablations.

\noindent
\textbf{Freeze vs Unfreeze Hiera encoder in HM2F-Video}: We investigate the performance obtained by the HM2F-Video framework while keeping its backbone frozen or fine-tuned, as shown in~\cref{tab:abl_frozen}. Keeping the Hiera encoder frozen results in an AuROC score of 77.79\%. However, fine-tuning the encoder during HM2F-Video training substantially improves performance, achieving an AuROC value of 93.44\% and an AuPRC value of 53.89\%.

\noindent\textbf{Number of frames}: We analyze the effect of varying the number of input frames in the CC-SAM2 method, as shown in~\Cref{tab:abl_frames}. We observe that the best performance is obtained when the number of frames is set to $2$, while the performance decreases drastically when the number of input frames increases to $8$. However, it is important to note that the frames are taken at 3 FPS (25 FPS original video annotated every 8th frame), resulting in weak correlation between consecutive frames. We hypothesize that a dataset with higher FPS would benefit from an increased number of input frames.

\section{Conclusion}
In this work, we introduced the road-obstacle video segmentation task. To evaluate the task: a) we curated four video road-segmentation datasets, b) evaluated nine image-based road-obstacle segmentation methods and two video semantic segmentation methods, along with our proposed baselines (CC-SAM2, and HM2F-Video), and c) selected evaluation metrics that assess model performance at pixel-, component-, and video-level. In addition, we conducted extensive evaluations to provide critical insight and discussion. We believe that road-obstacle video segmentation could pave the way for new tasks, such as few-shot road-obstacle video segmentation. 

\noindent\textbf{Acknowledgements.} This work is financially supported by the ELLIS Turin Unit. This work was partially funded by the ERC (853489 - DEXIM) and the Alfried Krupp von Bohlen und Halbach Foundation, which we thank for their generous support. The authors gratefully acknowledge the Gauss Centre for Supercomputing e.V.\ (\url{https://www.gauss-centre.eu}) for funding this project by providing computing time on the GCS Supercomputer JUWELS at Jülich Supercomputing Centre (JSC).

{
    \bibliographystyle{splncs04}
    \bibliography{main}

\begin{thebibliography}{10}
\providecommand{\url}[1]{\texttt{#1}}
\providecommand{\urlprefix}{URL }
\providecommand{\doi}[1]{https://doi.org/#1}

\bibitem{Maskomaly}
Ackermann, J., Sakaridis, C., Yu, F.: Maskomaly: Zero-shot mask anomaly segmentation. In: Brit. Mach. Vis. Conf. (2023)

\bibitem{Ahn_2024_ACCV}
Ahn, S., Jo, Y., Lee, K., Park, S.: Videopatchcore: An effective method to memorize normality for video anomaly detection. In: Proceedings of the Asian Conference on Computer Vision (ACCV). pp. 2179--2195 (December 2024)

\bibitem{bevandic2018discriminative}
Bevandi{\'c}, P., Kre{\v{s}}o, I., Or{\v{s}}i{\'c}, M., {\v{S}}egvi{\'c}, S.: Discriminative out-of-distribution detection for semantic segmentation. CoRR abs/1808.07703  (2018)

\bibitem{blum2021fishyscapes}
Blum, H., Sarlin, P.E., Nieto, J., Siegwart, R., Cadena, C.: The {Fishyscapes} benchmark: Measuring blind spots in semantic segmentation. Int. J. Comput. Vis.  \textbf{129}(11),  3119--3135 (2021)

\bibitem{chan2021segmentmeifyoucan}
Chan, R., Lis, K., Uhlemeyer, S., Blum, H., Honari, S., Siegwart, R., Salzmann, M., Fua, P., Rottmann, M.: {SegmentMeIfYouCan}: A benchmark for anomaly segmentation. Adv. Neural Inform. Process. Syst.  (2021)

\bibitem{chen2018encoder}
Chen, L.C., Zhu, Y., Papandreou, G., Schroff, F., Adam, H.: Encoder-decoder with atrous separable convolution for semantic image segmentation. In: Eur. Conf. Comput. Vis. p. 833–851 (2018)

\bibitem{M2F-Video}
Cheng, B., Choudhuri, A., Misra, I., Kirillov, A., Girdhar, R., Schwing, A.G.: Mask2former for video instance segmentation. CoRR  \textbf{abs/2112.10764} (2021), \url{https://arxiv.org/abs/2112.10764}

\bibitem{cheng2022masked}
Cheng, B., Misra, I., Schwing, A.G., Kirillov, A., Girdhar, R.: Masked-attention mask transformer for universal image segmentation. In: IEEE Conf. Comput. Vis. Pattern Recog. pp. 1280--1289 (2022)

\bibitem{mask2former}
Cheng, B., Misra, I., Schwing, A.G., Kirillov, A., Girdhar, R.: Masked-attention mask transformer for universal image segmentation. In: CVPR (2022)

\bibitem{cheng2021per}
Cheng, B., Schwing, A., Kirillov, A.: Per-pixel classification is not all you need for semantic segmentation. Adv. Neural Inform. Process. Syst.  (2021)

\bibitem{Cordts2016Cityscapes}
Cordts, M., Omran, M., Ramos, S., Rehfeld, T., Enzweiler, M., Benenson, R., Franke, U., Roth, S., Schiele, B.: The {Cityscapes} dataset for semantic urban scene understanding. In: IEEE Conf. Comput. Vis. Pattern Recog. pp. 3213--3223 (2016)

\bibitem{deng2009imagenet}
Deng, J., Dong, W., Socher, R., Li, L.J., Li, K., Fei-Fei, L.: {ImageNet}: A large-scale hierarchical image database. In: IEEE Conf. Comput. Vis. Pattern Recog. pp. 248--255 (2009)

\bibitem{di2021pixel}
Di~Biase, G., Blum, H., Siegwart, R., Cadena, C.: Pixel-wise anomaly detection in complex driving scenes. In: IEEE Conf. Comput. Vis. Pattern Recog. pp. 16913--16922 (2021)

\bibitem{DFN5B}
Fang, A., Jose, A.M., Jain, A., Schmidt, L., Toshev, A.T., Shankar, V.: Data filtering networks. In: The Twelfth International Conference on Learning Representations, {ICLR} 2024, Vienna, Austria, May 7-11, 2024. OpenReview.net (2024), \url{https://openreview.net/forum?id=KAk6ngZ09F}

\bibitem{fort2019deep}
Fort, S., Hu, H., Lakshminarayanan, B.: Deep ensembles: A loss landscape perspective. arXiv preprint arXiv:1912.02757  (2019)

\bibitem{gal2016dropout}
Gal, Y., Ghahramani, Z.: Dropout as a bayesian approximation: Representing model uncertainty in deep learning. In: Int. Conf. Mach. Learn. (2016)

\bibitem{grcic2020dense}
Grci{\'c}, M., Bevandi{\'c}, P., {\v{S}}egvi{\'c}, S.: Dense open-set recognition with synthetic outliers generated by real {NVP}. International Joint Conference on Computer Vision, Imaging and Computer Graphics Theory and Applications  (2021)

\bibitem{EAM}
Grcic, M., Saric, J., Segvic, S.: On advantages of mask-level recognition for outlier-aware segmentation. In: IEEE Conf. Comput. Vis. Pattern Recog. Worksh. pp. 2937--2947 (2023)

\bibitem{MAE}
He, K., Chen, X., Xie, S., Li, Y., Doll{\'{a}}r, P., Girshick, R.B.: Masked autoencoders are scalable vision learners. In: {IEEE/CVF} Conference on Computer Vision and Pattern Recognition, {CVPR} 2022, New Orleans, LA, USA, June 18-24, 2022. pp. 15979--15988. {IEEE} (2022). \doi{10.1109/CVPR52688.2022.01553}, \url{https://doi.org/10.1109/CVPR52688.2022.01553}

\bibitem{hendrycks2022improving}
Hendrycks, D., Basart, S., Mazeika, M., Zou, A., Kwon, J., Mostajabi, M., Steinhardt, J.: Improving and assessing anomaly detectors for large-scale settings  (2022)

\bibitem{maxlogit}
Hendrycks, D., Basart, S., Mazeika, M., Zou, A., Kwon, J., Mostajabi, M., Steinhardt, J., Song, D.: Scaling out-of-distribution detection for real-world settings. In: Chaudhuri, K., Jegelka, S., Song, L., Szepesv{\'{a}}ri, C., Niu, G., Sabato, S. (eds.) International Conference on Machine Learning, {ICML} 2022, 17-23 July 2022, Baltimore, Maryland, {USA}. Proceedings of Machine Learning Research, vol.~162, pp. 8759--8773. {PMLR} (2022), \url{https://proceedings.mlr.press/v162/hendrycks22a.html}

\bibitem{hendrycks2016baseline}
Hendrycks, D., Gimpel, K.: A baseline for detecting misclassified and out-of-distribution examples in neural networks. Int. Conf. Learn. Represent.  (2017)

\bibitem{ApolloScape}
Huang, X., Wang, P., Cheng, X., Zhou, D., Geng, Q., Yang, R.: The apolloscape open dataset for autonomous driving and its application. {IEEE} Trans. Pattern Anal. Mach. Intell.  \textbf{42}(10),  2702--2719 (2020). \doi{10.1109/TPAMI.2019.2926463}, \url{https://doi.org/10.1109/TPAMI.2019.2926463}

\bibitem{jung2021standardized}
Jung, S., Lee, J., Gwak, D., Choi, S., Choo, J.: Standardized max logits: A simple yet effective approach for identifying unexpected road obstacles in urban-scene segmentation. In: Int. Conf. Comput. Vis. pp. 15405--15414 (2021)

\bibitem{CSVPS}
Kim, D., Woo, S., Lee, J., Kweon, I.S.: Video panoptic segmentation. In: 2020 {IEEE/CVF} Conference on Computer Vision and Pattern Recognition, {CVPR} 2020, Seattle, WA, USA, June 13-19, 2020. pp. 9856--9865. Computer Vision Foundation / {IEEE} (2020)

\bibitem{sam}
Kirillov, A., Mintun, E., Ravi, N., Mao, H., Rolland, C., Gustafson, L., Xiao, T., Whitehead, S., Berg, A.C., Lo, W.Y., Doll{\'a}r, P., Girshick, R.: Segment anything. In: ICCV (2023)

\bibitem{lin2017refinenet}
Lin, G., Milan, A., Shen, C., Reid, I.: {RefineNet}: Multi-path refinement networks for high-resolution semantic segmentation. In: IEEE Conf. Comput. Vis. Pattern Recog. pp. 5168--5177 (2017)

\bibitem{lin2014microsoft}
Lin, T.Y., Maire, M., Belongie, S., Hays, J., Perona, P., Ramanan, D., Doll{\'a}r, P., Zitnick, C.L.: Microsoft {COCO}: Common objects in context. In: Eur. Conf. Comput. Vis. p. 740–755 (2014)

\bibitem{lis2019detecting}
Lis, K., Nakka, K., Fua, P., Salzmann, M.: Detecting the unexpected via image resynthesis. In: Int. Conf. Comput. Vis. pp. 2152--2161 (2019)

\bibitem{liu2020energy}
Liu, W., Wang, X., Owens, J., Li, Y.: Energy-based out-of-distribution detection. Adv. Neural Inform. Process. Syst.  (2020)

\bibitem{liu2021swin}
Liu, Z., Lin, Y., Cao, Y., Hu, H., Wei, Y., Zhang, Z., Lin, S., Guo, B.: Swin transformer: Hierarchical vision transformer using shifted windows. In: ICCV (2021)

\bibitem{long2015fully}
Long, J., Shelhamer, E., Darrell, T.: Fully convolutional networks for semantic segmentation. In: IEEE Conf. Comput. Vis. Pattern Recog. pp. 3431--3440 (2015)

\bibitem{loshchilov2017decoupled}
Loshchilov, I., Hutter, F.: Decoupled weight decay regularization. Int. Conf. Learn. Represent.  (2019)

\bibitem{SOS}
Maag, K., Chan, R., Uhlemeyer, S., Kowol, K., Gottschalk, H.: Two video data sets for tracking and retrieval of out of distribution objects. In: Wang, L., Gall, J., Chin, T., Sato, I., Chellappa, R. (eds.) Computer Vision - {ACCV} 2022 - 16th Asian Conference on Computer Vision, Macao, China, December 4-8, 2022, Proceedings, Part {V}. Lecture Notes in Computer Science, vol. 13845, pp. 476--494. Springer (2022). \doi{10.1007/978-3-031-26348-4\_28}, \url{https://doi.org/10.1007/978-3-031-26348-4\_28}

\bibitem{VC}
Miao, J., Wei, Y., Wu, Y., Liang, C., Li, G., Yang, Y.: {VSPW:} {A} large-scale dataset for video scene parsing in the wild. In: {IEEE} Conference on Computer Vision and Pattern Recognition, {CVPR} 2021, virtual, June 19-25, 2021. pp. 4133--4143. Computer Vision Foundation / {IEEE} (2021). \doi{10.1109/CVPR46437.2021.00412}, \url{https://openaccess.thecvf.com/content/CVPR2021/html/Miao\_VSPW\_A\_Large-scale\_Dataset\_for\_Video\_Scene\_Parsing\_in\_the\_CVPR\_2021\_paper.html}

\bibitem{mukhoti2018evaluating}
Mukhoti, J., Gal, Y.: Evaluating bayesian deep learning methods for semantic segmentation. CoRR abs/1811.12709  (2018)

\bibitem{RbA}
Nayal, N., Yavuz, M., Henriques, J.F., G{\"{u}}ney, F.: {RbA}: Segmenting unknown regions rejected by all. In: Int. Conf. Comput. Vis. pp. 711--722 (2023)

\bibitem{lostandfound}
Pinggera, P., Ramos, S., Gehrig, S., Franke, U., Rother, C., Mester, R.: Lost and found: detecting small road hazards for self-driving vehicles. In: 2016 {IEEE/RSJ} International Conference on Intelligent Robots and Systems, {IROS} 2016, Daejeon, South Korea, October 9-14, 2016. pp. 1099--1106. {IEEE} (2016). \doi{10.1109/IROS.2016.7759186}, \url{https://doi.org/10.1109/IROS.2016.7759186}

\bibitem{clip}
Radford, A., Kim, J.W., Hallacy, C., Ramesh, A., Goh, G., Agarwal, S., Sastry, G., Askell, A., Mishkin, P., Clark, J., et~al.: Learning transferable visual models from natural language supervision. In: ICML (2021)

\bibitem{mask2anomaly}
Rai, S.N., Cermelli, F., Caputo, B., Masone, C.: Mask2anomaly: Mask transformer for universal open-set segmentation. T-PAMI  (2024)

\bibitem{rai2023unmasking}
Rai, S.N., Cermelli, F., Fontanel, D., Masone, C., Caputo, B.: Unmasking anomalies in road-scene segmentation. Int. Conf. Comput. Vis. pp. 4014--4023 (2023)

\bibitem{Ramachandra-WACV-2020}
Ramachandra, B., Jones, M.J.: Street scene: A new dataset and evaluation protocol for video anomaly detection. In: 2020 IEEE Winter Conference on Applications of Computer Vision (WACV). pp. 2558--2567 (2020)

\bibitem{sam2}
Ravi, N., Gabeur, V., Hu, Y.T., Hu, R., Ryali, C., Ma, T., Khedr, H., R{\"a}dle, R., Rolland, C., Gustafson, L., Mintun, E., Pan, J., Alwala, K.V., Carion, N., Wu, C.Y., Girshick, R., Doll{\'a}r, P., Feichtenhofer, C.: Sam 2: Segment anything in images and videos. arXiv preprint arXiv:2408.00714  (2024), \url{https://arxiv.org/abs/2408.00714}

\bibitem{Ristea-CVPR-2024}
Ristea, N.C., Croitoru, F.A., Ionescu, R.T., Popescu, M., Khan, F.S., Shah, M.: Self-distilled masked auto-encoders are efficient video anomaly detectors. In: 2024 IEEE/CVF Conference on Computer Vision and Pattern Recognition (CVPR). pp. 15984--15995 (2024)

\bibitem{Hiera}
Ryali, C., Hu, Y., Bolya, D., Wei, C., Fan, H., Huang, P., Aggarwal, V., Chowdhury, A., Poursaeed, O., Hoffman, J., Malik, J., Li, Y., Feichtenhofer, C.: Hiera: {A} hierarchical vision transformer without the bells-and-whistles. In: Krause, A., Brunskill, E., Cho, K., Engelhardt, B., Sabato, S., Scarlett, J. (eds.) International Conference on Machine Learning, {ICML} 2023, 23-29 July 2023, Honolulu, Hawaii, {USA}. Proceedings of Machine Learning Research, vol.~202, pp. 29441--29454. {PMLR} (2023), \url{https://proceedings.mlr.press/v202/ryali23a.html}

\bibitem{LiderSOD}
Singh, A., Kamireddypalli, A., Gandhi, V., Krishna, K.M.: Lidar guided small obstacle segmentation. In: {IEEE/RSJ} International Conference on Intelligent Robots and Systems, {IROS} 2020, Las Vegas, NV, USA, October 24, 2020 - January 24, 2021. pp. 8513--8520. {IEEE} (2020). \doi{10.1109/IROS45743.2020.9341465}, \url{https://doi.org/10.1109/IROS45743.2020.9341465}

\bibitem{tian2022pixel}
Tian, Y., Liu, Y., Pang, G., Liu, F., Chen, Y., Carneiro, G.: Pixel-wise energy-biased abstention learning for anomaly segmentation on complex urban driving scenes. In: Eur. Conf. Comput. Vis. p. 246–263 (2022)

\bibitem{vaswani2017attention}
Vaswani, A., Shazeer, N., Parmar, N., Uszkoreit, J., Jones, L., Gomez, A.N., Kaiser, {\L}., Polosukhin, I.: Attention is all you need. Adv. Neural Inform. Process. Syst.  (2017)

\bibitem{vojir2021road}
Vojir, T., {\v{S}}ipka, T., Aljundi, R., Chumerin, N., Reino, D.O., Matas, J.: Road anomaly detection by partial image reconstruction with segmentation coupling. In: Int. Conf. Comput. Vis. pp. 15631--15640 (2021)

\bibitem{xia2020synthesize}
Xia, Y., Zhang, Y., Liu, F., Shen, W., Yuille, A.: Synthesize then compare: Detecting failures and anomalies for semantic segmentation. In: Eur. Conf. Comput. Vis. p. 145–161 (2020)

\bibitem{ZhangSBHL23}
Zhang, D., Sakmann, K., Beluch, W., Hutmacher, R., Li, Y.: Anomaly-aware semantic segmentation via style-aligned ood augmentation. In: Int. Conf. Comput. Vis. Worksh. pp. 4067--4075 (2023)

\bibitem{DVIS}
Zhang, T., Tian, X., Wu, Y., Ji, S., Wang, X., Zhang, Y., Wan, P.: {DVIS:} decoupled video instance segmentation framework. In: Int. Conf. Comput. Vis. pp. 1282--1291 (2023)

\bibitem{zhang2018exfuse}
Zhang, Z., Zhang, X., Peng, C., Xue, X., Sun, J.: {ExFuse}: Enhancing feature fusion for semantic segmentation. In: Eur. Conf. Comput. Vis. p. 273–288 (2018)

\bibitem{zhou2017scene}
Zhou, B., Zhao, H., Puig, X., Fidler, S., Barriuso, A., Torralba, A.: Scene parsing through {ADE20K} dataset. In: IEEE Conf. Comput. Vis. Pattern Recog. pp. 5122--5130 (2017)

\bibitem{zhu2020deformable}
Zhu, X., Su, W., Lu, L., Li, B., Wang, X., Dai, J.: Deformable {DETR}: Deformable transformers for end-to-end object detection. Int. Conf. Learn. Represent.  (2021)

\end{thebibliography}
}

\begin{center}
    {\LARGE \textbf{Road Obstacle Video Segmentation}} \\[1ex] 
    {\large (Supplementary Material)} 
\end{center}

In the following sections, we present additional analysis~(\Cref{sec:analysis}), along with extended dataset details in~\Cref{sec:dataset}.
\begin{figure}[h!]
  \centering
   \includegraphics[width=0.9\linewidth, height=0.8\linewidth]{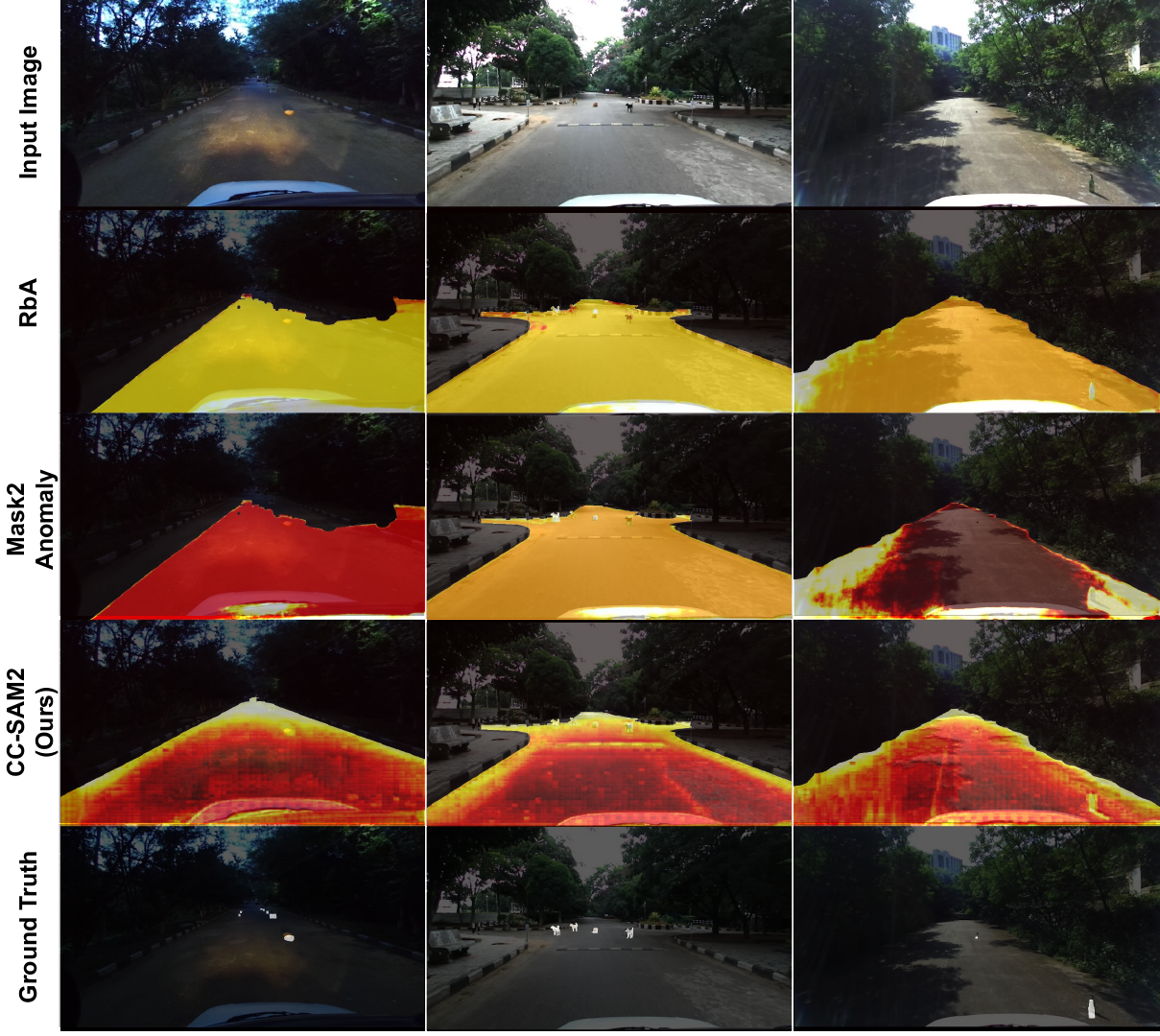}
   \caption{
    Predicted road-obstacle maps obtained from the baselines (RbA~\cite{RbA}, Mask2Anomaly~\cite{mask2anomaly}) and our proposed method CC-SAM2 on the LidarSOD dataset~\cite{LiderSOD}.}
   \label{fig:supp_cvpr_fig3}
\end{figure}
\begin{figure}[t]
  \centering
   \includegraphics[width=0.99\linewidth]{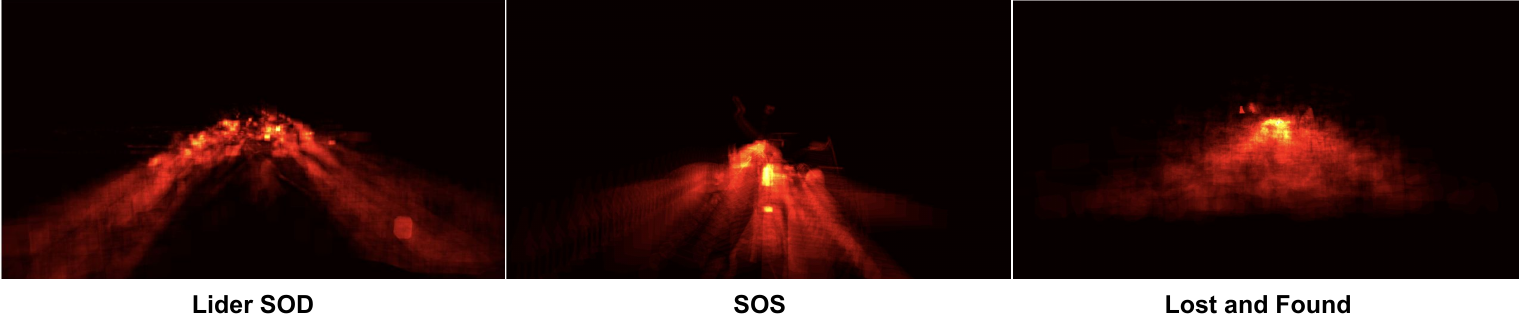}
   \caption{Average road-obstacles pixel distribution in different datasets. This illustrates the average location of the obstacles on the road.} 
   \label{fig:supp_4}
\end{figure}
\begin{figure*}[t]
  \centering
   \includegraphics[width=0.99\linewidth]{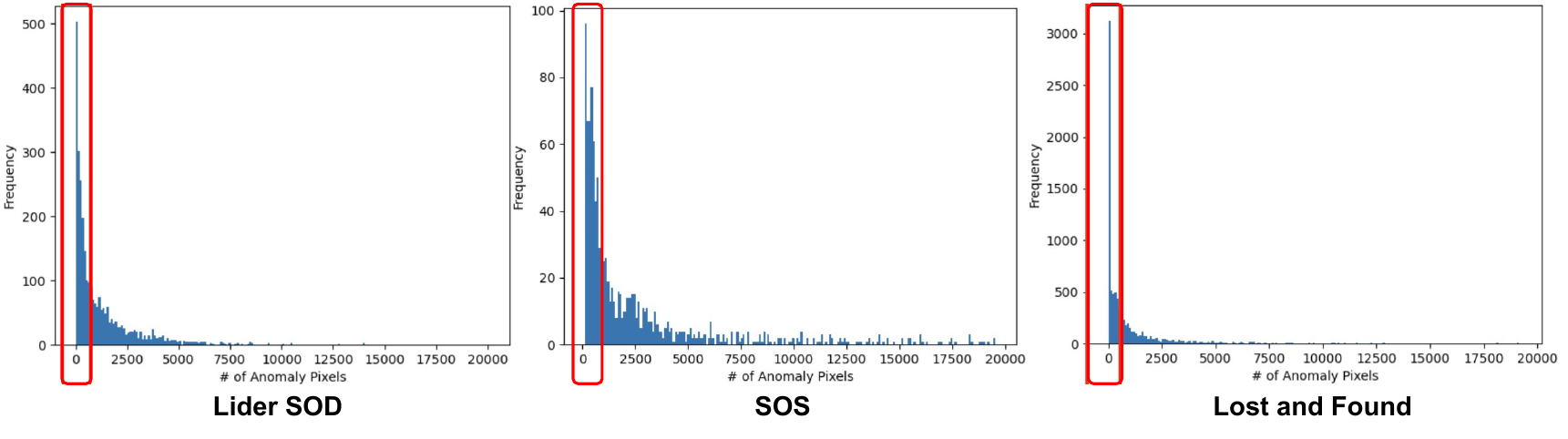}
   \caption{Shows the frequency of road obstacles having varying sizes. Inside the red bounding boxes, it shows that the majority of road obstacles are small in size.} 
   \label{fig:supp_5}
\end{figure*}
\begin{figure}[t]
  \centering
   \includegraphics[width=0.99\linewidth]{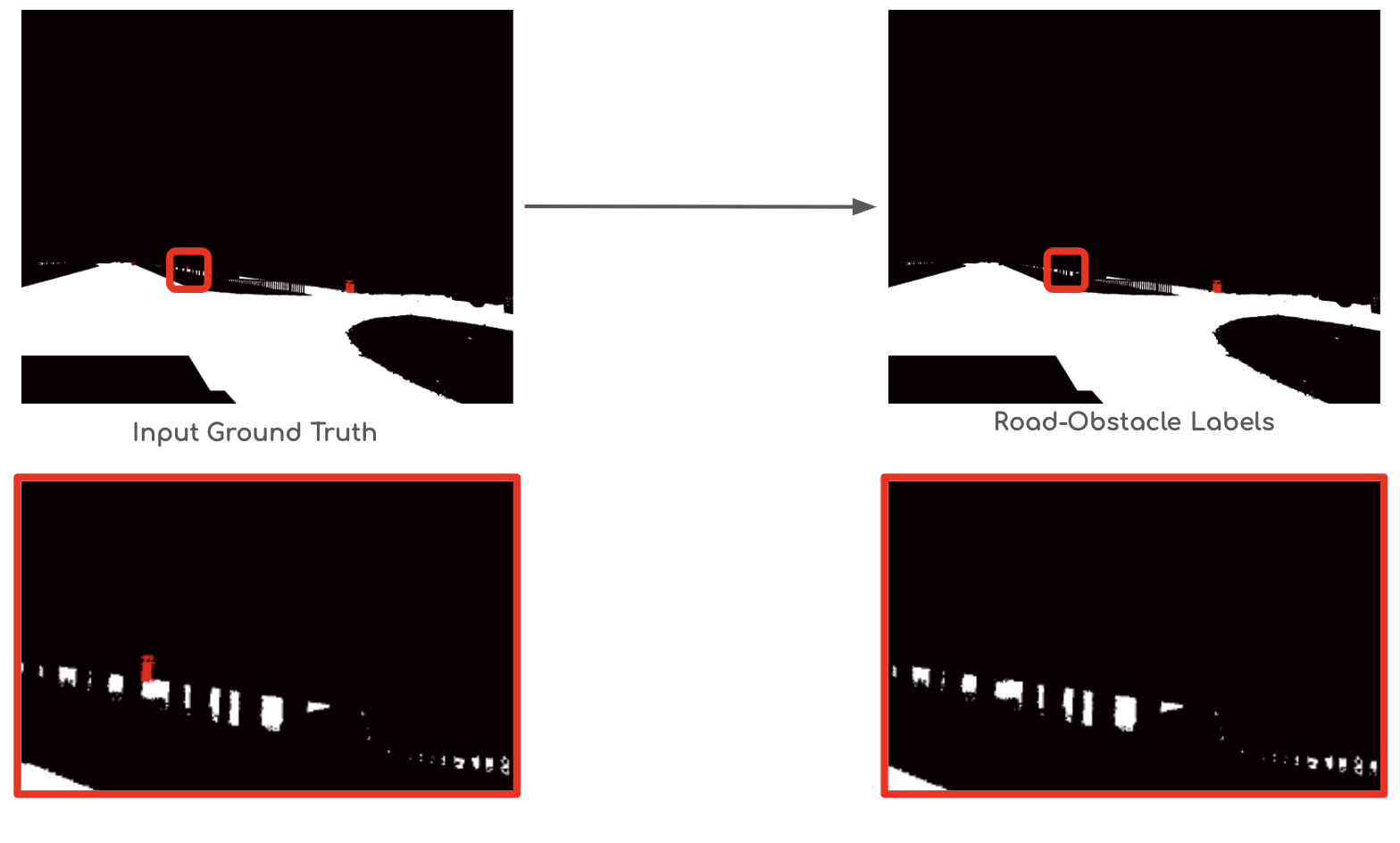}
   \caption{We show that while curating Apolloscape-Road-Obstacle, the intersection of road segment with the obstacle centroids help removing the obstacles that are outside the road.} 
   \label{fig:supp_6}
\end{figure}

\section{Additional Analysis}\label{sec:analysis}
\subsection{Domain shift Effect}
In this section, we discuss the effect of the domain shift in the road-obstacle video segmentation performance. The LidarSOD~\cite{LiderSOD} dataset is collected from an Indian road scene scenario, which is significantly different from the training dataset Cityscapes-VPS in terms of traffic scenarios and scene layout, since the latter is captured in Germany. As a result, we qualitatively observe in~\cref{fig:supp_cvpr_fig3}, that all methods predict many false positives. In addition,~\cref{fig:supp_cvpr_fig3} first column shows the input image captured during the night, where we can observe that under such poor illumination conditions the false positives are increased further. Thus, making road-obstacle segmentation more challenging.

\subsection{Qualitative Examples}
\begin{figure*}[t]
  \centering
   \includegraphics[width=0.99\linewidth]{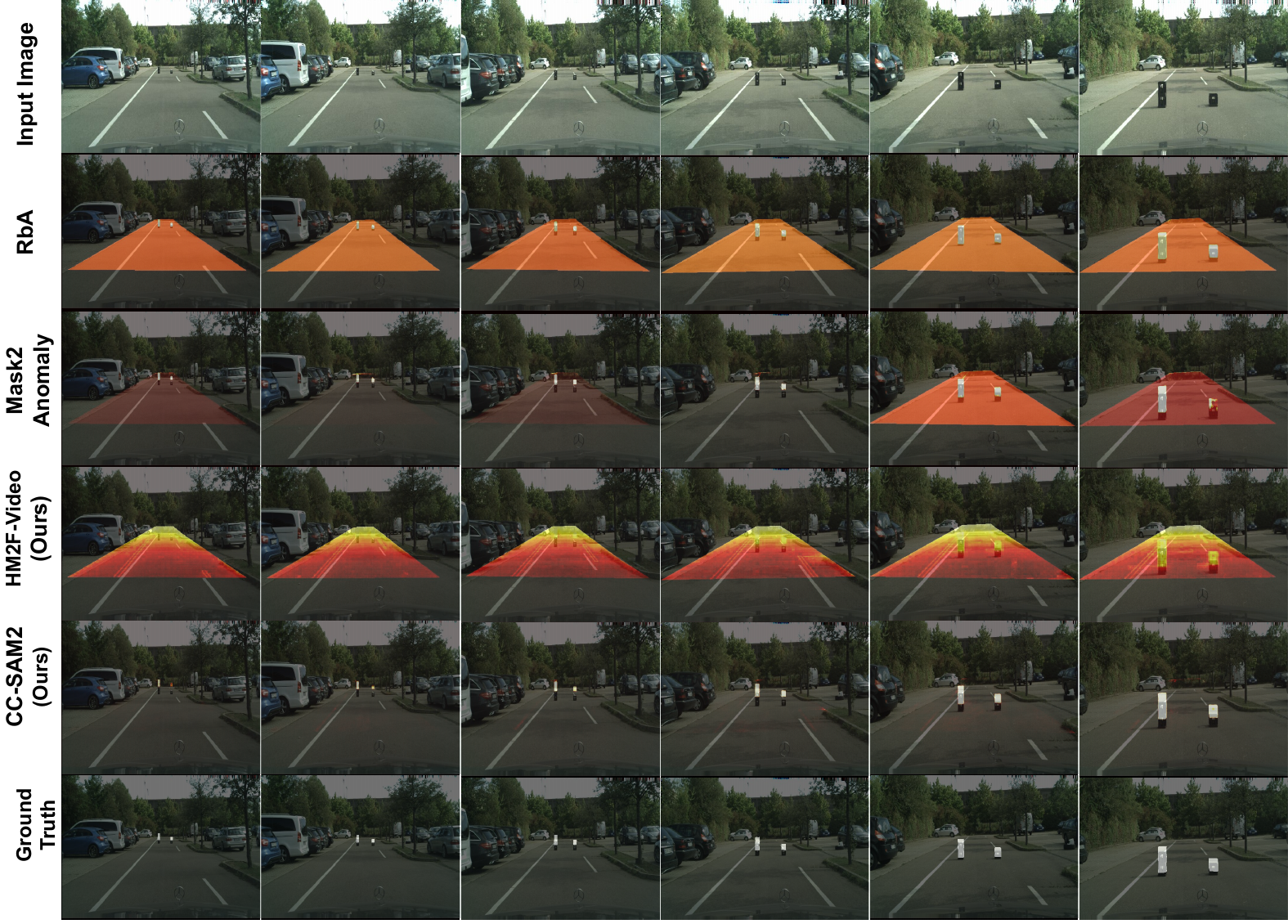}
   \caption{Predicted road-obstacle maps obtained from the baselines (RbA~\cite{RbA}, Mask2Anomaly~\cite{mask2anomaly}) and our proposed method CC-SAM2 and HM2F-Video on the Lost\&Found dataset~\cite{lostandfound}.} 
   \label{fig:supp_cvpr_fig1}
\end{figure*}

\begin{figure*}[t]
  \centering
   \includegraphics[width=0.99\linewidth]{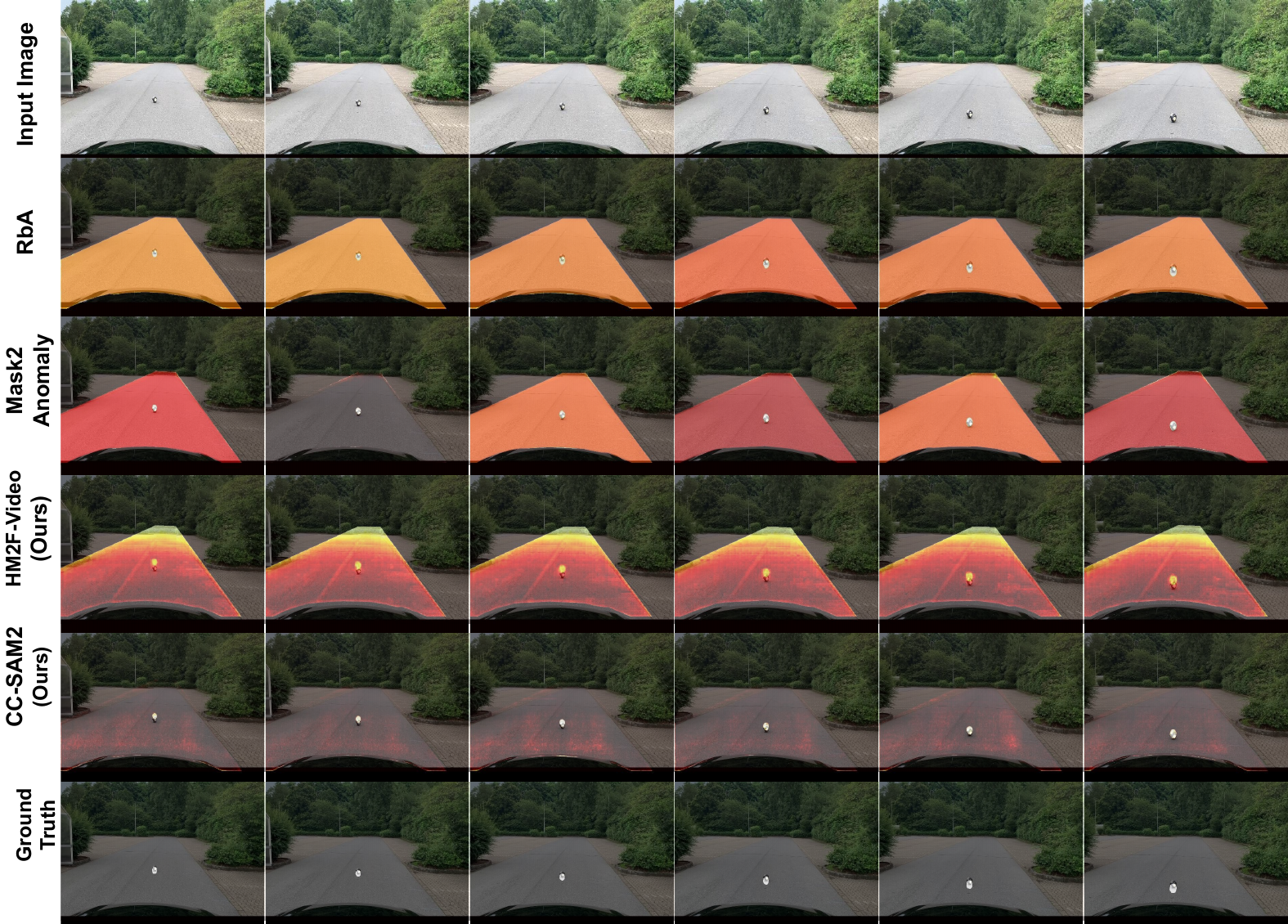}
   \caption{Predicted road-obstacle maps obtained from the baselines (RbA~\cite{RbA}, Mask2Anomaly~\cite{mask2anomaly}) and our proposed method CC-SAM2 and HM2F-Video on the SOS dataset~\cite{SOS}.} 
   \label{fig:supp_cvpr_fig2}
\end{figure*}

We provide further qualitative examples in Figs.~\ref{fig:supp_cvpr_fig1} and \ref{fig:supp_cvpr_fig2}. Compared to prior image-based methods (\eg RbA~\cite{RbA}, Mask2Anomaly~\cite{mask2anomaly}), we observe that CC-SAM2 provides far more accurate segmentation maps, further underscoring the benefits of utilizing video foundation models for our task. We also notice that the masks generated by CC-SAM2 are more accurate than those of HM2F-Video.

\section{Additional Dataset Details} \label{sec:dataset}
\textbf{Threshold on road-obstacle size: }\Cref{fig:supp_5} illustrates the distribution of road obstacles of various sizes across different datasets. Notably, very small road obstacles appear frequently in all datasets. To address this, we applied a size threshold, removing obstacles smaller than 100 pixels in LidarSOD and 225 pixels in both SOS and Lost\&Found. \\
\textbf{Average road-obstacle pixel distribution: }
In~\cref{fig:supp_4}, we present the average pixel distribution of road obstacles across different datasets. In Lost\&Found~\cite{lostandfound} and SOS~\cite{SOS}, obstacles are primarily located in the middle of the road, whereas in LidarSOD, they tend to appear along the road's periphery.

\noindent
\textbf{Apolloscape-Road-Obstacle dataset curation details:} While curating the AsRO dataset, some obstacles were not present on the road, but had an overlap with some road pixels, as shown in~\cref{fig:supp_6} (inside zoomed box). Therefore, we must remove such obstacles, since they are not on the road. We first calculate the convex hull of all obstacle instances. Then, for each obstacle instance, we calculate the intersection of the road segment with the obstacle centroid. Following the aforementioned steps, we remove such road obstacles as shown in~\cref{fig:supp_6}.\\
\noindent\textbf{SOS~\cite{SOS}: } The SOS dataset comprises 20 video sequences captured in real world road environments. Each sequence is recorded at 25 FPS, with every eighth frame annotated.  This results in a total of 1,129 pixel-level semantically-annotated images. The obstacles present in the dataset follow the same labeling method as the Cityscapes dataset.  It includes 13 types of obstacles present in the road scenes, such as bags and balls. Additionally, the SOS dataset contains two classes: road-obstacle/out-of-distribution (OOD) and Road. 

All other areas apart from OOD and road are considered void.

\noindent
\textbf{LidarSOD~\cite{LiderSOD}:} comprises 15 video sequences. The total number of annotated images is 2919. The ground truth maps have three classes: road, off-road (buildings, cars, pedestrians etc.), and obstacle (with nine types of classes like stones, bricks, dogs etc.). \\
\textbf{Lost\&Found~\cite{lostandfound}:} consists of 13 different street scene sequences with 37 different types of obstacles that are present on the road. The dataset provides coarse annotations of free-space areas and pixel-level fine-grained annotations of obstacles present on the road. The dataset has a total of 2239 annotated frames. 
\section{Implementation Details}
All the baseline models are based on the mask-transformers~\cite{mask2former,M2F-Video} architecture. During inference, the model outputs a set of \textit{class masks} $M$ along with their corresponding \textit{class scores} $C$, which are used to compute anomaly scores. In our work, we establish two baseline categories:(a) Image-based obstacle segmentation methods, and (b) Video-based semantic segmentation methods. We provide a detailed implementation discussion of each category and corresponding methods in the following sub-sections.
\subsection{Image-based obstacle segmentation methods}
All the image obstacle segmentation methods are based on Mask2Former~\cite{mask2former}, which consists of an image encoder, a pixel decoder, and a transformer decoder. We used Swin-B as the image encoder, and its weights are initialized from pre-trained ImageNet~\cite{deng2009imagenet}. Next, we utilized a multi-scale deformable attention transformer~\cite{zhu2020deformable} as the pixel decoder, giving feature maps at $1/8, 1/16,$ and $1/32$ resolution to provide image features to the transformer decoder layers. The transformer decoder consists of 9 layers (except for RbA~\cite{RbA} where the number of decoder layers is 1) with 100 queries. 

The training details for the image-based obstacle segmentation are mainly derived from~\cite{cheng2021per,cheng2022masked}. We train each of the methods on the Cityscapes~\cite{Cordts2016Cityscapes} dataset, with an initial learning rate of 1e-4 and batch size of 16 for 90 thousand iterations on AdamW~\cite{loshchilov2017decoupled} with a weight decay of 0.05. We now describe how to obtain anomaly scores for each image-based obstacle segmentation method.
\begin{itemize}
    \item \textbf{Maximum softmax}~\cite{hendrycks2016baseline} is the most commonly used obstacle segmentation method. Given an image $I$, the anomaly score $S_{MSP}$ can be calculated as:
    \begin{equation*}
        S_{MSP}(I) = 1 - \max (C^T \times M).
    \end{equation*}
    \item \textbf{Energy}~\cite{liu2020energy} detect obstacle based on the difference of energies between in-distribution (known classes) and out-of-distribution (obstacles). We calculate anomaly score for an image $I$ as: 
    \begin{equation*}
        S_{Energy}(I) = -logsumexp(C^T \times M).
    \end{equation*}
    \item \textbf{Entropy}~\cite{hendrycks2016baseline} in anomaly segmentation measures the uncertainty. With higher entropy indicating anomaly or road obstacle region. Now, for an image $I$, we can calculate the anomaly score as:
    \begin{equation*}
        S_{Entropy}(I) = -p\text{log}p,\:\:	p = (C^T \times M)
    \end{equation*} 
    \item \textbf{Max Logit}~\cite{maxlogit} have shown to greatly outperforms the maximum softmax. Max logit can be calculated as negative of the maximum unnormalized logit to get anomaly score that can be equated as: 
        \begin{equation*}
        S_{Max Logit}(I) = - \max (C^T \times M)
    \end{equation*} 
    \item \textbf{Void Classifier}~\cite{blum2021fishyscapes} utilizes the background class as anomaly or road obstacle class. The anomaly score can be yielded as: 
    \begin{equation*}
        S_{Void Classifier}(I) = (C^T \times M)_{V}
    \end{equation*} 
    $V$ here represents the score obtained from the void class.
\end{itemize}

The remaining image obstacle segmentation methods: Mask2Anomaly~\cite{rai2023unmasking}, EAM~\cite{EAM}, AEM~\cite{EAM}, and RbA~\cite{RbA} utilize an additional fine-tuning step called outlier exposure, where synthetic anomalies are utilized to further improve the road-obstacle segmentation performance. Now, we will discuss each of the method in more detail.

\begin{itemize}
    \item~\textbf{Mask2Anomaly~\cite{mask2anomaly}} during outlier exposure train the model using mask-contrastive loss. The synthetic outlier image is generated using AnomalyMix~\cite{tian2022pixel} where an object from MS-COCO~\cite{lin2014microsoft} dataset image is cut and pasted on the Cityscapes image. The network is trained for 4000 iterations with a learning rate of 1e-5, and batch size 8, keeping all the other hyper-parameters the same. The probability of choosing an outlier image in a training batch is kept at 0.2. During inference the anomaly score is calculated using maximum softmax.

    \item~\textbf{RbA~\cite{RbA}} performs outlier exposure by sampling 300 images from MS-COCO. The model is trained for 2K iterations with a batch size of 16. The probability of choosing an outlier image in a training batch is set to 0.2. To calculate anomaly score during inference, it employs the following equation:
    \begin{equation}
        S_{RbA}(I) = -\sum \gamma(C^T \times M).
    \end{equation}
    $\gamma$ represents the $tanh$ function.

    \item~\textbf{EAM~\cite{EAM} and AEM~\cite{EAM}} fine-tune its model for 2K iterations during the outlier exposure step utilizing the  ADE20K~\cite{zhou2017scene} to generate the outlier images. AEM during anomaly inference use max logit to get anomaly scores whereas EAM employs the following equation to detect anomalous region: 
    \begin{equation*}
        S_{EAM}(I) = -  (C^T \times \max (M))
    \end{equation*} 
    
\end{itemize}

\subsection{Video-based semantic segmentation methods}
Since no video road obstacle segmentation method previously existed, we repurposed existing video-based semantic segmentation approaches: M2F-Video~\cite{M2F-Video} and DVIS~\cite{DVIS}. Both methods are based on Mask2Former~\cite{mask2former}, which we adapted for our task by applying the Max Softmax technique~\cite{hendrycks2016baseline} to get anomaly scores. We train both methods on the Cityscapes-VPS dataset~\cite{CSVPS}. In the rest of the section, we provide a detailed implementation detail of each approach.

\begin{itemize}
    \item \textbf{M2F-Video~\cite{M2F-Video}} is trained using the AdamW optimizer with an initial learning rate of 0.0001 and a weight decay of 0.05. The training is conducted for 16 thousand iterations with a batch size of 4. During training, we sample two consecutive frames from each video clip and resize them to 512\(\times\)256 pixels.
    \item \textbf{DVIS~\cite{DVIS}} comprises of M2F-Video~\cite{M2F-Video} and a temporal refiner consisting of six temporal decoder blocks. It is trained using the AdamW optimizer with a base learning rate of 1e-4 and a weight decay of 5e-2. The training is performed for 16,000 iterations with a batch size of 4.
    
\end{itemize}

\section{Zero-shot Road-Obstacle Video Segmentation}

SAM 2 gives class-agnostic outputs that limits its utility for road-obstacle segmentation task. To address this, we use CLIP’s~\cite{clip} zero-shot classification capabilities. We inject the known class information from the class taxonomy defined in Cityccapes~\cite{Cordts2016Cityscapes}, \eg ``road'', ``building'', ``wall'', ``traffic light'', as these classes are commonly encountered in urban road scenes. 

Given an input video frame at time $t$ as $I^{t}$, we initially pass it through the pre-trained SAM 2 model. We utilize a grid of equally sampled points as the segmentation prompt for SAM 2. The output of SAM 2 is a set of masks $\mathcal{M}= \{m_1, m_2, \dots, m_n\}$, where $n$ is the number of masks. For each mask $m_i$, $0 \le i \le n$, we crop the content of the bounding box that encloses the segment delimitated by the mask $m_i$. Subsequently, we classify the crop content of $m_i$ with CLIP, which gives a set of class probabilities $P_i = \{p_1, p_2, \dots, p_j\}$, where $j$ is the number of text prompts. Next, we assign the road-obstacle segmentation score for the segment $m_i$ as: $
s_i = 1 - \max(P_i) $. Finally, we aggregate all masks along with their scores to get prediction $\hat{A}^{t}$ for $I^{t}$.
\begin{equation}
    s_i = 1 - \max(P_i)
\end{equation}

We term the entire pipeline as Zero-Shot SAM2 (ZS-SAM2). ZS-SAM2 employs CLIP~\cite{clip} with ViT-H trained in DFN-5B~\cite{DFN5B} and SAM2 architecture with the hiera large backbone~\cite{Hiera}. From~\cref{tab:supp1}, we observe that ZS-SAM2 performs significantly poor than CC-SAM2, despite being supervised by CLIP. This indicates that while SAM2 has demonstrated impressive zero-shot performance across various computer vision tasks, it struggles with more complex scenarios such as road-obstacle video segmentation. This suggests that further modifications are necessary to adapt off-the-shelf foundation models for more complex tasks such as road-obstacle video segmentation. 

\begin{table*}[t]
\centering
\setlength{\tabcolsep}{10pt}
\caption{Comparison between the zero-shot performance of SAM2 (ZS-SAM2) and its fine-tuned variant (CC-SAM2).}

\resizebox{0.99\linewidth}{!}{\begin{tabular}{@{}llccccccc@{}}
\toprule
& \multirow{2}{*}{\bf Methods} & \multicolumn{3}{c}{\bf Pixel-Level Metrics} & \multicolumn{3}{c}{\bf Component-Level Metrics} & \multicolumn{1}{c}{\bf Video Metrics} \\
\cmidrule(lr){3-5} \cmidrule(lr){6-8} \cmidrule(lr){9-9} 
\textbf{Dataset}& & AuROC$\uparrow$ & AuPRC$\uparrow$ & FPR$_{95}\downarrow$ & sIoU$\uparrow$ & PPV$\uparrow$ & F1$^{*}_{1}\uparrow$ & VC$^{*}\uparrow$\\

\midrule
Lost\&Found\\
\addlinespace[0.5em]
\midrule
& ZS-SAM2	&53.5	&2.1	&99.7	&18.2	&9.6	&9.6	&0.0 \\
& CC-SAM2 	&98.4	&57.8	&6.2	&15.1	&38.9	&14.7	&54.2  \\
\addlinespace[0.5em]

\midrule
SOS\\
\addlinespace[0.5em]
\midrule
& ZS-SAM2	&76.0	&2.7	&73.7	&17.8	&4.2	&5.7	&0.0\\
& CC-SAM2 	&99.2	&79.2	&3.7	&31.1	&47.0	&28.1	&43.3  \\
\addlinespace[0.5em]

\midrule
Lidar SOD\\
\addlinespace[0.5em]
\midrule
& ZS-SAM2	&62.8	&0.4	&96.1	&2.4	&6.2	&1.6	&0.0\\
& CC-SAM2 	&92.5	&14.0	&25.6	&9.1	&5.0	&3.2	&40.5  \\
\addlinespace[0.5em]

\bottomrule

\end{tabular}}
\label{tab:supp1}
\end{table*}

\section{Limitation and Future Direction}
One of the limitations of our benchmarks is the sparse annotations, since the most densely annotated dataset is SOS with 3 FPS. A more densely annotated dataset would further improve segmentation precision.
Furthermore, the evaluation of all methods is conducted using seven metrics across three categories, which leads to a fragmented assessment and makes direct method comparison difficult. Introducing a unified ranking scheme or incorporating metric-importance weighting would significantly enhance interpretability of an method performance on the task.
\end{document}